# One-Shot Learning from Demonstration of Contact-Rich Robotic Manipulation by Identifying Physical Interactions

A.H.G. Overbeek, H. van der Kooij, and M. Vlutters

***Abstract*— Learning from Demonstration (LfD) allows robots to learn manipulation tasks directly from humans, thereby supporting the versatile application of robots. Most LfD methods do not explicitly model the physical interactions between a robot and its environment, such as the making and breaking of contact, while these are crucial during manipulation tasks. Because the same basic physical interactions recur often, they can be a basis for robust, generalizable, and adaptive task reproduction. We propose an LfD method that explicitly uses *what* physical interactions take place *where* and *when*. Using that information, a hybrid position-force controller tracks demonstrated trajectories until contact-based transition conditions from the demonstrations are met. We evaluate our method in real robot experiments consisting of opening doors and locks, bolt picking and screwing, dislodging, and surface contouring. We show that explicitly modeling physical interactions benefits LfD in four ways. First, by allowing reproduction of complex, sequential, and contact-rich manipulation tasks using only a single demonstration and no prior knowledge of the task. Second, by facilitating robustness to unknown geometric variations in the environment. Third, by facilitating generalization when geometric variations are known. Fourth, by facilitating online adaptation using geometric information explored during task reproduction. We discuss how robustness, generalization, and adaptivity can be explicitly implemented, which is generally lacking in the LfD literature. Thereby, our work aims to close a gap in interpretable few-shot LfD of robotic manipulation.**

## I. INTRODUCTION

Autonomous robots can assist humans within various labor settings, such as in healthcare, logistics, and the manufacturing industry. Robots may be required to perform many different manipulation tasks, such as pick-and-place tasks, contact tasks such as cleaning and assembly, and tool-related tasks such as machining. These may require several differently configured robots, because a particular robot may lack the versatility to complete a range of different tasks. Therefore, more versatile robots that are capable of various manipulation tasks may be more cost-effective.

### *A. Robot Learning*

Machine learning may improve the versatile application of robots by reducing the need for manual programming. Such methods may be split into two sequential steps. First, imitating an expert by inferring control directly from demonstrated data, referred to as Learning from Demonstration (LfD), imitation learning, or behavioral cloning [1]. Second, iteratively improving over time by optimizing a predefined measure of task success, such as a reward function in reinforcement learning [2]. Alternatively, in inverse reinforcement learning, a reward function may be learned from demonstration data to avoid defining it manually [3].

Although various general-purpose learning methods have been applied to robotic manipulation, they may require more physical data (kinematics, forces) than is readily available, particularly when compared to the availability of representational data (visual, language) available on the internet. Learning manipulation may require a substantial amount of physical data, because complex, contact-rich, and sequential manipulation tasks involve long time horizons with multi-dimensional positions, velocities, and forces related by nonlinear dynamics and constraints for each handled object. In addition, manipulation tasks may contain known or unknown variations, such as the properties of the objects and the environment.

The limited availability of physical data may be addressed in several ways. First, more data can be collected in the real world or in simulation. Simulated data may be substantially cheaper, faster, and safer to collect, but additional effort is required to transfer the learned outcome to the real world [4]. Second, available real-world data from multiple sources may be accumulated into large public datasets, but such data may be of varying quality and completeness [5], [6], [7], [8]. For example, haptic information such as interaction forces essential to contact-rich tasks may be lacking. Such tasks are common during manufacturing and (dis)assembly, such as polishing, insertion, and screwing. Furthermore, relying on general-purpose learning methods trained on large datasets generally results in large models that are difficult to interpret, debug, extend, and provide performance and safety guarantees for. Moreover, these approaches require substantial computational resources.

Alternatively, the need for data may be reduced by developing more efficient learning methods specific to robotics. This work focuses on such robotics-specific LfD methods, using the domain knowledge available in robotics research. We consider the imitation learning step (LfD), because improvements there may propagate to subsequent improvement steps, such as reinforcement learning.

### B. Related Work

To reduce the need for real-world data, robotics-specific LfD methods have been proposed that require only a few demonstrations and little task-specific prior knowledge.

#### 1) Few-Shot Learning from Demonstration for Robotics

The majority of few-shot LfD methods split task modeling into two levels [9].

Trajectory models describe the instantaneous properties of signals, such as poses, velocities, forces, and impedances. They are commonly learned with trajectory descriptors such as Probabilistic Movement Primitives [10], Gaussian Processes [11], or Gaussian Mixture Models [12], [13], [14]. Such models capture trajectories and their variability between demonstrations, and provide generalization to new geometric situations. Alternatively, the demonstrated trajectories can be modeled as dynamical systems, for example with Dynamic Movement Primitives [15] or Stable Estimated Dynamics Systems [16].

High-level models describe (sub)tasks with constant properties, such as goals. They are often learned with Hidden Markov Models that cluster common positions, velocities, and forces [14], [17], [18], [19]. These methods typically do not capture physically meaningful states, such as contact sequences, or capture them implicitly. Instead, they rely on clustering general commonalities between demonstrations.

#### 2) Explicitly Modeling Physical Interactions

In contrast, explicitly modeling the physical interactions that occur in a demonstration has several advantages for few-shot LfD [20]. We split physical interactions into three aspects: *what*, *when*, and *where* (Fig. 1).

First, based on *what* interactions occur *when*, specific control methods can be chosen at particular moments over general-purpose methods. For example, position, velocity, or force control may be chosen over impedance control with constant stiffness. This can improve performance while keeping interaction forces low [21]. Second, meaningful (sub) task segments can be derived from the identified sequence of interactions. Such task segments may recur in many different manipulation tasks. They can be used to recognize similar tasks, and generalize the reproduction to similar geometric situations, for example using vision [22], [23], [24], [25]. Third, planning for specific interactions can improve robust task reproduction. For example, intentionally introducing contacts can simplify alignment tasks [26]. Alternatively, avoiding contacts can simplify reaching tasks [27].

In addition, tasks may be simplified based on *where* tasks are expressed. In particular, signals may become small and decoupled from each other in geometric features that correspond to constraints, such as contact points [21]. Because reference frames are essential in robotics, placing them in such relevant geometric features has been proposed as a basis for task modeling [21], [28]. Reference frames may be selected from several options that are defined beforehand [29], [30], [31], [32]. Alternatively, reference frames may be identified without prior information in a decision process [33] or through optimization [20], [34].

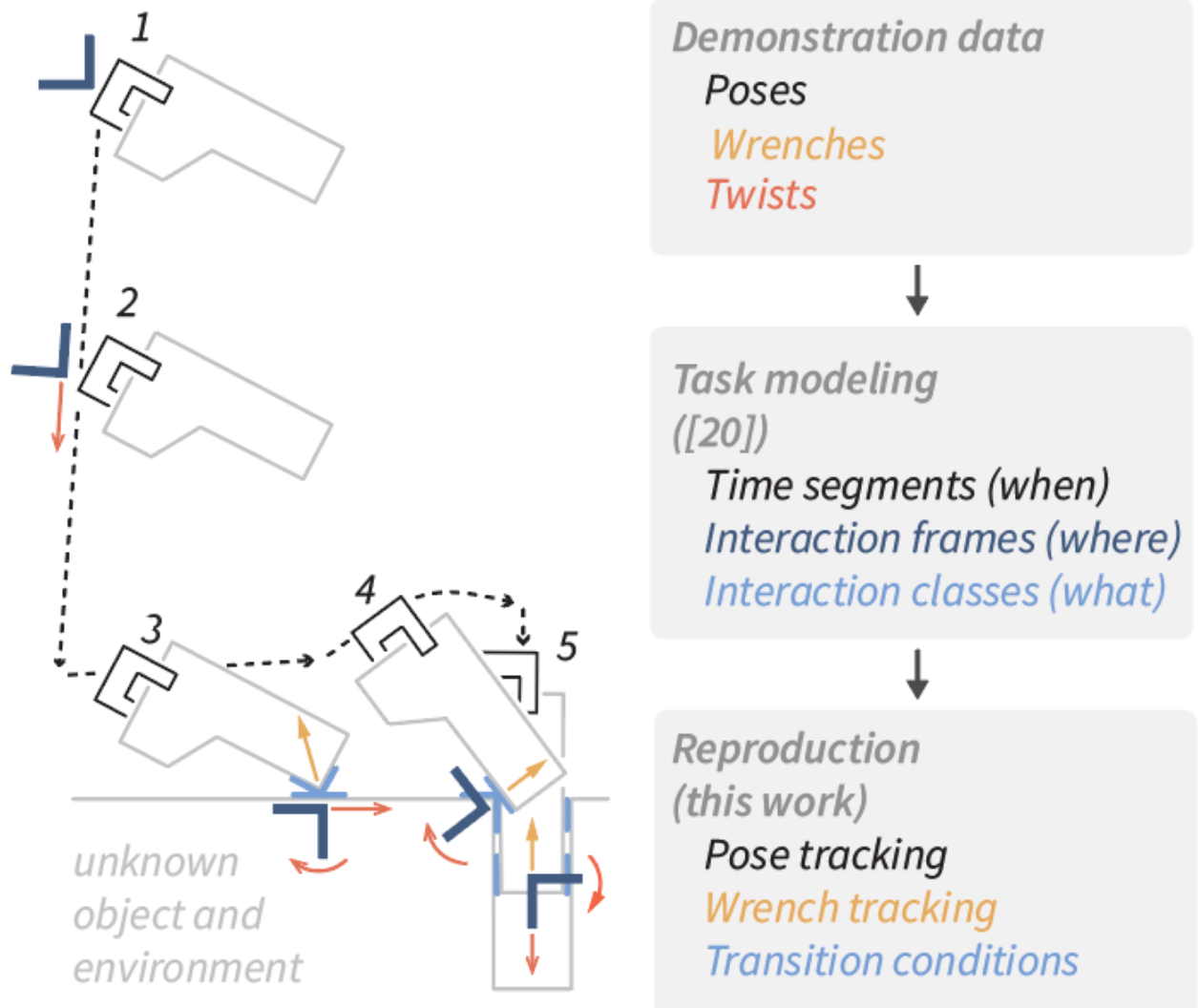

**Fig. 1.** An example of a peg-in-hole task demonstration, consisting of idle waiting (**1**), free-space approaching (**2**), partially constrained hole search (**3**), partially constrained alignment (**4**), and highly constrained insertion (**5**). Using only the recorded poses, twist, and wrenches, our previous work identified *what* physical interactions occurred *when* and *where*, such as at meaningful geometric features [20]. We extend that work by deriving contact-aware transition conditions and by tracking demonstrated poses and wrenches. Thereby, we reproduce complex, forceful, and sequential manipulation tasks from a single demonstration without any task-specific prior information, such as object and environment geometry.

### C. Contribution and Outline

Related work (Section I.B.2) has illustrated that explicitly modeling aspects of physical interactions (*what, where, when*) can support LfD in several ways. However, they have not simultaneously modeled all three for the LfD of complex, sequential, and contact-rich tasks.

We propose LfD following identification of those basic physical interactions, by expanding our previous work on contact-based task modeling (Fig. 1, [20]). That work identified *what* interactions (interaction classes) took place *where* (interaction frame) and *when* (segments) in a single task demonstration. We expand that method in two ways. First, by deriving transition conditions between segments (*how to transition*, Section III). Second, by facilitating force and trajectory tracking and generalization thereof (*how to track*, Section IV and V).

We apply conventional reproduction control (Section V) to illustrate four advantages of explicitly modeling physical interactions for LfD. First, by facilitating learning of complex, contact-rich, and sequential tasks using only a single demonstration and no task-specific prior knowledge. Second, by facilitating robustness to unknown geometric variations through interaction-aware control. Third, by facilitating generalization to new situations given a priori geometric information, such as from a vision system. Fourth, by facilitating online adaptation following exploration, such as trajectory extrapolation.

Because our method is human-interpretable, we explicitly discuss how robustness, generalization, and adaptivity may be implemented using only a single demonstration. Such insights may be difficult to obtain for approaches with limited interpretability, such as black box methods.

We evaluate our method in several robot tasks: opening locked cabinets, bolt picking and screwing, and surface following (Section VI).

## II. Preliminaries

### A. Assumptions

Manipulation tasks may consist of complex physical interactions. We make several assumptions to limit our scope. First, the environment is assumed to be static, apart from rigidly grasped objects. Second, the grasped objects and environment are rigid. Third, gravity is compensated for. Fourth, manipulation is quasi-static. We therefore ignore structural compliance and inertia, but contacts may include static and kinetic friction.

### B. Rigid Body Transformations

The motion between two rigid bodies can be represented by the transformation between two reference frames, one frame rigidly attached to each body. Such reference frames have an origin and an orientation in space, which can change with body motion. The transformation between two reference frames in three-dimensional space can be parameterized by a distance between their origins $\boldsymbol{o} \in \mathbb{R}^3$ and a rotation matrix $\boldsymbol{R} \in \mathrm{SO}(3)$ between their orientations [35]. These components can be combined in a homogenous *transformation matrix*

$$\boldsymbol{T} = \begin{bmatrix} \boldsymbol{R} & \boldsymbol{o} \\ \boldsymbol{0} & 1 \end{bmatrix} \quad \in \mathrm{SE}(3),$$

which represents the pose of one frame relative to the other. Alternatively, the pose can be stored in a *pose tuple*

$$\boldsymbol{P} = (\boldsymbol{R}, \boldsymbol{o}) \quad \in \mathrm{SO}(3) \times \mathbb{R}^3.$$

The orientation $\boldsymbol{R}$ can also be represented by a rotation vector $\boldsymbol{\theta} \in \mathbb{R}^3$, related by the exponential map

$$\boldsymbol{R} = \exp([\boldsymbol{\theta}]_\times) \quad \in \mathrm{SO}(3),$$

where $[\cdot]_\times$ denotes the skew-symmetric representation

$$[\boldsymbol{\theta}]_\times = \begin{bmatrix} 0 & -\theta_3 & \theta_2 \\ \theta_3 & 0 & -\theta_1 \\ -\theta_2 & \theta_1 & 0 \end{bmatrix} \quad \in \mathrm{so}(3).$$

The inverse skew-symmetric operation is denoted with $[\cdot]^\vee$ such that $[\ln(\boldsymbol{R})]^\vee = \boldsymbol{\theta}$.

The instantaneous velocities and forces between two rigid bodies can be represented by a twist $\boldsymbol{t}$ and a wrench $\boldsymbol{w}$:

$$\boldsymbol{t} = \begin{bmatrix} \boldsymbol{\omega} \\ \boldsymbol{v} \end{bmatrix} \quad \in \mathbb{R}^6, \qquad \boldsymbol{w} = \begin{bmatrix} \boldsymbol{m} \\ \boldsymbol{f} \end{bmatrix} \quad \in \mathbb{R}^6,$$

which contain angular velocities $\boldsymbol{\omega} \in \mathbb{R}^3$, linear velocities $\boldsymbol{v} \in \mathbb{R}^3$, moments $\boldsymbol{m} \in \mathbb{R}^3$ and forces $\boldsymbol{f} \in \mathbb{R}^3$.

Twists and wrenches are expressed in a reference frame denoted by a superscript, for example $\boldsymbol{t}^b$ is a twist expressed in reference frame $\{b\}$.

If $\boldsymbol{T}$ describes the pose of a body frame $\{b\}$ with respect to a ground frame $\{g\}$, the twist $\boldsymbol{t}^b = [\boldsymbol{\omega}^b, \boldsymbol{v}^b]$ of the body with respect to the ground expressed in $\{b\}$ is related to $\boldsymbol{T}$ by:

$$\begin{bmatrix} [\boldsymbol{\omega}^b]_\times & \boldsymbol{v}^b \\ \boldsymbol{0} & 0 \end{bmatrix} = \boldsymbol{T}^{-1}\dot{\boldsymbol{T}} \quad \in \mathrm{se}(3),$$

where $\dot{\boldsymbol{T}}$ is the time derivative of $\boldsymbol{T}$.

Expressing the twist $\boldsymbol{t}^b$ or wrench $\boldsymbol{w}^b$ in a new frame other than $\{b\}$ requires the pose matrix of $\{b\}$ with respect to the new frame. For example, if $\boldsymbol{R}$ and $\boldsymbol{o}$ again represent the pose of body frame $\{b\}$ with respect to the ground frame $\{g\}$, the transformations

$$\boldsymbol{t}^g = \begin{bmatrix} \boldsymbol{R} & \boldsymbol{0} \\ [\boldsymbol{o}]_\times \boldsymbol{R} & \boldsymbol{R} \end{bmatrix} \boldsymbol{t}^b, \qquad \boldsymbol{w}^g = \begin{bmatrix} \boldsymbol{R} & [\boldsymbol{o}]_\times \boldsymbol{R} \\ \boldsymbol{0} & \boldsymbol{R} \end{bmatrix} \boldsymbol{w}^b \tag{1}$$

express the twist and wrench in frame $\{g\}$. We refer to the *twist-wrench pair* with

$$\boldsymbol{s} = (\boldsymbol{t}, \boldsymbol{w}) \quad \in \mathbb{R}^{12},$$

and to the *pose-wrench* pair with

$$\boldsymbol{x} = (\boldsymbol{P}, \boldsymbol{w}) \quad \in \mathrm{SO}(3) \times \mathbb{R}^9.$$

## III. Extended Task Modeling

This section summarizes previous our previous work on task modeling (Section III.A, [20]) and extends it with the derivation of transition conditions between segments (Sections III.B and III.C).

### A. Task Modeling Summary of [20]

Recording a task demonstration yields poses, twists, and wrenches. They are recorded in the pose-wrench pair $\boldsymbol{x}^b(k)$ and twist-wrench pair $\boldsymbol{s}^b(k)$ expressed in the end effector frame $\{b\}$, where $k$ is discrete time. From that data, our previous work identified *what* physical interactions (interaction classes $\boldsymbol{S}$) occured *where* (interaction frame $\{\chi\}$) and *when* ($N$ time segments) (Fig. 1, Table I).

#### 1) Time Segmentation (When)

After recording a task, various segmentation methods can be used. In our previous work, data was segmented where the norms of the signals ($\|\cdot\| \in \mathbb{R}_+$) in the twist-wrench pair $\boldsymbol{s}^b(k)$ crossed prespecified thresholds ($^*\omega, ^*v, ^*m, ^*f \in \mathbb{R}_+$). Subsequently, extraneous segments were filtered out. In this work, we make several filtering improvements (Appendix A).

TABLE I
IDENTIFIED TASK MODEL PARAMETERS

| Component | Symbol | Domain |
|---|---|---|
| Segment (*when*) | $n$ | $\{0,1,\dots,N\}$ |
| Interaction frame (*where*) | $\{_n\chi\}$<br>$_n\boldsymbol{R}^\chi$<br>$_n\boldsymbol{o}^\chi$ | $\{\mathcal{GG},\mathcal{BB},\mathcal{GB},\mathcal{BG}\}$<br>$SO(3)$<br>$\mathbb{R}^3$ |
| Interaction class (*what*) | $_n\boldsymbol{\mathcal{S}}^\chi$ | $\{\mathcal{I},\mathcal{F},\mathcal{Z},\mathcal{C}\}^{12}$ |
| Pose-wrench pair *(how to track)* | $_n\boldsymbol{x}^\chi(\tau)$ | $SO(3)\times\mathbb{R}^9$ |
| Transition conditions *(how to transition)* | $_n\boldsymbol{\pi}$ | $\{^\wedge c, {}^\vee c, {}^>\tau\}^{12}$ |

We therefore obtain $N$ segments, denote the $n$-th segment with the left subscript, and normalize discrete time $k$ to a progress variable $\tau \in [0,1]$ for each segment. For example, $_2\boldsymbol{v}^b(\tau)$ is the linear velocity of the end effector during the second segment, expressed in the end effector.

*2) Interaction Frame Identification (Where)*

After segmenting a task, our previous work identified a task-relevant reference frame (interaction frame $\{_n\chi\}$) for each segment $n$ to express the task in. This was done by identifying frames which simultaneously minimized and decoupled the cartesian mechanical power components in the demonstration. Thereby, simultaneously minimizing and decoupling the twists and wrenches.

Each frame is of one of four types $\{_n\chi\} \in \{\mathcal{GG},\mathcal{BB},\mathcal{GB},\mathcal{BG}\}$, where the first letter (e.g. $\mathcal{G}$ for ground) denotes in which body the frame orientation is constant, and the second letter (e.g. $\mathcal{B}$ for end effector body) in which body the frame origin is constant. Each segment's twist-wrench pair $_n\boldsymbol{s}^b(\tau)$ is expressed in that segment's interaction frame, denoted $_n\boldsymbol{s}^\chi(\tau)$ (1).

*3) Interaction Classification (What)*

After identifying the interaction frames $\{_n\chi\}$, our previous work determined the interaction classes $\boldsymbol{\mathcal{S}} = \{\mathcal{I},\mathcal{F},\mathcal{Z},\mathcal{C}\}^6$ in each of those frames, consisting of three rotations and three translations about the three axes of a frame. The classes were:

$\mathcal{I}$: Idle (no substantial wrenches or twists),
$\mathcal{C}$: Constrained (substantial wrenches, but no twists),
$\mathcal{F}$: Free motion (substantial twists, but no wrenches),
$\mathcal{Z}$: Impeded motion (substantial twists and wrenches).

The classes $\boldsymbol{\mathcal{S}}$ were determined by applying thresholds $\varsigma(\bar{\boldsymbol{s}}) = \boldsymbol{\mathcal{S}}$, where $\bar{\boldsymbol{s}} \in \mathbb{R}_+^{12}$ are the root mean squares (RMS) over time of the twist-wrench pair $\boldsymbol{s}(\tau)$. For example, if $\bar{\omega}_1 < {}^*\omega$ and $\bar{m}_1 > {}^*m$, rotations about the first axis were considered constrained ($\mathcal{C}$).

When modeling six degrees of freedom with six interaction classes, all interactions (e.g. constraints) were assumed to be bilateral.

## B. *Deriving Unilateral Interaction Classes*

To model unilateral interactions, such as when contacting a plane, we extend our previous work such that the positive and negative directions of frame axes may have different interaction classes. This results in twelve interaction classes instead of six. To that end, we split the twist-wrench pair $\boldsymbol{s}(\tau) \in \mathbb{R}^{12}$ into components corresponding to the positive and negative directions of each frame axis (${}^+\boldsymbol{s}(\tau) \in \mathbb{R}_+^{12}$, ${}^-\boldsymbol{s}(\tau) \in \mathbb{R}_-^{12}$)

$$\boldsymbol{s}(\tau) = {}^+\boldsymbol{s}(\tau) + {}^-\boldsymbol{s}(\tau).$$

We do so based on whether the twist-wrench pair lies in the top-right ($+$) or bottom-left ($-$) half-region of the force-velocity plot, determined by the dashed black line in Fig. 2b. We then separately classify the positive and negative directions

$${}^+\boldsymbol{\mathcal{S}} = \varsigma({}^+\bar{\boldsymbol{s}}), \qquad {}^-\boldsymbol{\mathcal{S}} = \varsigma({}^-\bar{\boldsymbol{s}}),$$

to obtain twelve interaction classes $\boldsymbol{\mathcal{S}} = ({}^+\boldsymbol{\mathcal{S}}, {}^-\boldsymbol{\mathcal{S}}) \in \{\mathcal{I},\mathcal{F},\mathcal{Z},\mathcal{C}\}^{12}$ (Fig. 2, Table I). For example, ${}^{+\text{rot}}_{\ \ 2}\mathcal{S}_1^\chi = \mathcal{C}$ indicates that the positive rotation ($+$rot) around the first axis of a frame ($\{\chi\}$) is constrained during the second segment.

For static environments, the classes of an axis' positive and negative directions must be either the same (e.g. ${}^\pm\mathcal{S} = \mathcal{F}$), or one is idle (e.g. ${}^+\mathcal{S} = \mathcal{F}$ and ${}^-\mathcal{S} = \mathcal{I}$). A counterexample is a ratchet, which allows movement in one direction but is constrained in the other (${}^+\mathcal{S} = \mathcal{F}$ and ${}^-\mathcal{S} = \mathcal{C}$). Because such a mechanism requires a third moving part, the environment is not considered completely static.

## C. *Inferring Transition Conditions*

To infer what transitions ($\rightarrow$) took place in a demonstration, we observe how the interaction classes $\boldsymbol{\mathcal{S}} \in \{\mathcal{I},\mathcal{F},\mathcal{Z},\mathcal{C}\}^{12}$ change from one segment to the next. For example, ${}^{+\text{lin}}_{\ \ 3}\mathcal{S}_1^\chi = \mathcal{F} \rightarrow \mathcal{C}$ indicates that moving along the positive direction of the first axis of frame $\{\chi\}$ resulted in a constraint at the end of the third segment. In this work, we infer the following simple transition conditions with symbols (Fig. 2b):

${}^\wedge c$: Constraint making, when $\mathcal{S} \in \{\mathcal{F},\mathcal{Z}\} \rightarrow \mathcal{C}$.
${}^\vee c$: Constraint breaking, when $\mathcal{S} = \mathcal{C} \rightarrow \mathcal{F},\mathcal{Z}$.
${}^>\tau$: Tracking progress completion, for all other cases.

During ${}^>\tau$, we assume that a human demonstrator tracked certain unknown reference signals until some *measured* desired state was reached. That is, all information necessary to control transitions between segments is assumed to be present in the measured data.

Demonstration in geometric space

(a)

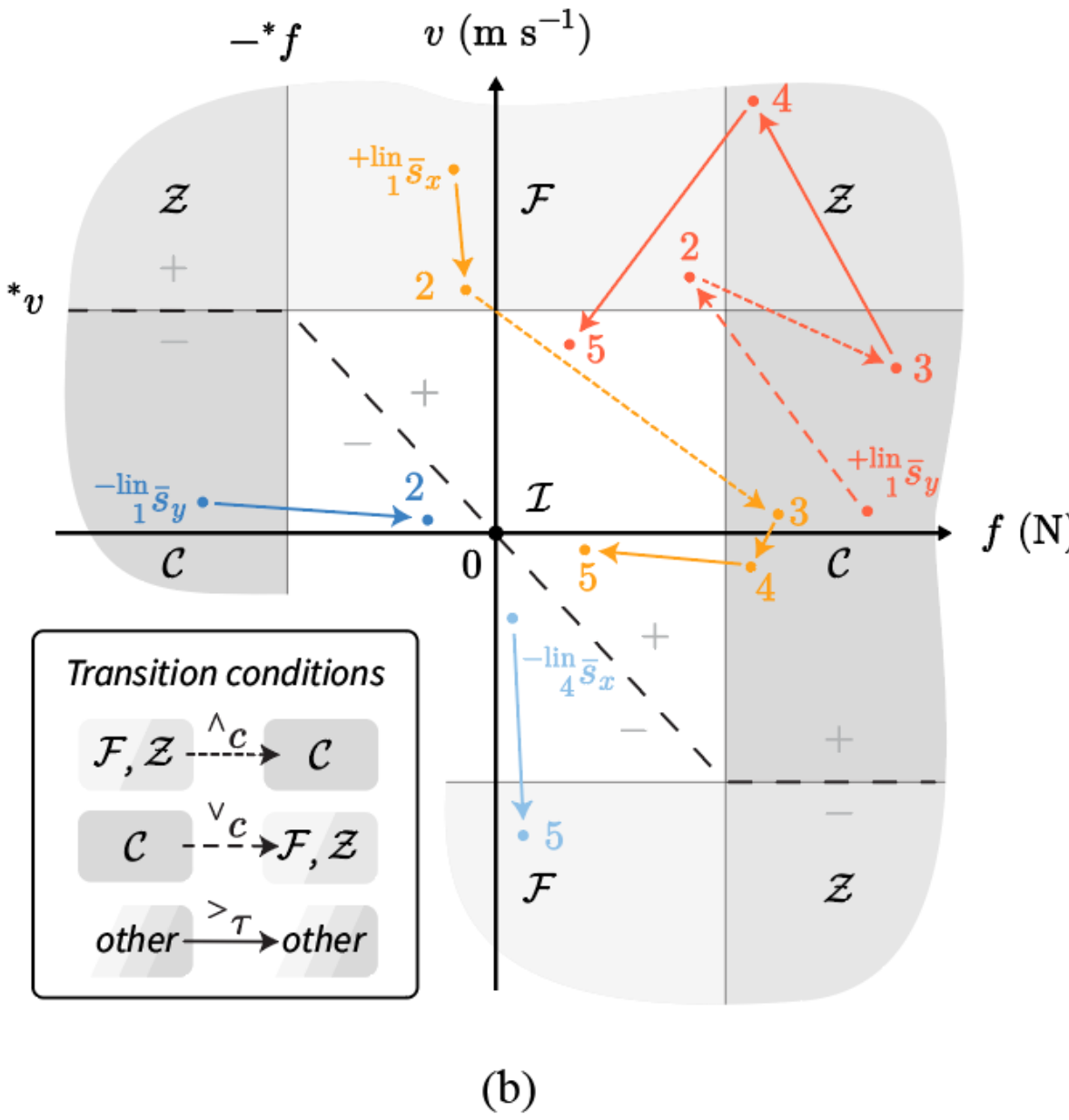

(b)

**Fig. 2.** Deriving the interaction classes and transition conditions of an example 5-segment task demonstration. This example is purely planar and translational. (a) A demonstrator (**1**) moves a frictionless box through a slit until it loses a constraint (${}^{\vee}c$) off the edge, (**2**) moves through free space until contact is made (${}^{\wedge}c$), (**3**) pushes to overcome static friction (${}^{\vee}c$), (**4**) pushes through kinetic friction to stop at a chosen position, and (**5**) lifts the box. (b) For all $n$ segments, we split the twist-wrench pair ${}^{\text{lin}}_{n}\boldsymbol{s}(\tau)$ into ${}^{+\text{lin}}_{n}\boldsymbol{s}(\tau)$ and ${}^{-\text{lin}}_{n}\boldsymbol{s}(\tau)$ and determine their RMS ${}^{\pm\text{lin}}_{n}\bar{\boldsymbol{s}}$. Those RMS may become small, but not zero due to noise. We then apply thresholds (${}^{*}v$ and ${}^{*}f$ ) to each segment's ${}^{\pm\text{lin}}_{n}\bar{\boldsymbol{s}}$ to obtain the interaction classes ${}^{\text{lin}}_{n}\boldsymbol{S} \in \{\mathcal{I},\mathcal{F},\mathcal{Z},\mathcal{C}\}^4$ indicated by the shaded regions. Based on the class transitions, we infer the transition conditions ${}^{\text{lin}}_{n}\boldsymbol{\pi} \in \{{}^{\wedge}c, {}^{\vee}c, {}^{>}\tau\}^4$ with constraint making (${}^{\wedge}c$ when $\mathcal{F},\mathcal{Z} \to \mathcal{C}$), constraint breaking (${}^{\vee}c$ when $\mathcal{C} \to \mathcal{F},\mathcal{Z}$), and tracking completion (${}^{>}\tau$ for all other cases). In general, demonstrations are non-planar and also contain rotations, and so we determine twelve instead of four classes and transitions per segment, and each segment has its own interaction frame.

We denote the transition conditions of segment $n \to n+1$ as

$$ {}_{n}\boldsymbol{\pi} = \{{}^{\wedge}c, {}^{\vee}c, {}^{>}\tau\}^{12}. $$

This results in twelve conditions for each segment: one for each rotation about, and translation along both the positive and negative axes of each three-dimensional interaction frame. For example, ${}^{+\text{lin}}_{3}\boldsymbol{\pi}^{\chi}_{1} = {}^{\wedge}c$ indicates that a linear constraint is encountered along the positive direction of the first axis of frame $\{\chi\}$ at the end of the third segment. During reproduction, these transition conditions will be monitored to control switching to the next segment.

The transitions between two segments can only be compared if they are expressed in the same interaction frame. Therefore, the twist-wrench pair of segments ${}_{n}\boldsymbol{s}(\tau)$ and ${}_{n+1}\boldsymbol{s}(\tau)$ can be expressed in either segment's frame $\{{}_{n}\chi\}$ or $\{{}_{n+1}\chi\}$:

$$ \begin{aligned} \varsigma({}_{n}\bar{\boldsymbol{s}}^{\,n\chi}) &\to \varsigma({}_{n+1}\bar{\boldsymbol{s}}^{\,n\chi}), \\ \varsigma({}_{n}\bar{\boldsymbol{s}}^{\,n+1\chi}) &\to \varsigma({}_{n+1}\bar{\boldsymbol{s}}^{\,n+1\chi}). \end{aligned} $$

For each transition, we choose to use the frame $\{{}_{n}\chi\}$ or $\{{}_{n+1}\chi\}$ that contains the lowest sum of ${}^{\wedge}c$ and ${}^{\vee}c$ conditions. That frame may be the sparser representation, because fewer conditions must be monitored for the same transition.

## IV. Pose Trajectories and Geometric Generalization

This section describes how pose trajectories can be parameterized and generalized using interaction frames. From the task demonstration data, we obtain poses, twists and wrenches for each segment. To reproduce the trajectories of a task, we track the demonstrated poses, because tracking twists may lead to drift with respect to the desired final pose.

### A. Interaction Frame Trajectories

Instead of tracking the end effector pose with respect to the ground, the changes in the identified interaction frames may be tracked. Our previous work identified interaction frames with a constant orientation $\boldsymbol{R}^{\chi}$ in the ground *or* body, and a constant displacement $\boldsymbol{o}^{\chi}$ in the ground *or* body [20]. This results in four frame types (Table I). For example, a frame of type $\{\mathcal{GB}\}$ has a constant orientation $\boldsymbol{R}^{\chi}$ in the *ground* frame and a constant displacement $\boldsymbol{o}^{\chi}$ in the *body* frame. Conversely, that frame has a time-dependent orientation $\widetilde{\boldsymbol{R}}^{\chi}(\tau)$ in the body frame, and a time-dependent displacement $\widetilde{\boldsymbol{o}}^{\chi}(\tau)$ in the ground frame. This holds analogously for the other three frame types.

To track the demonstrated trajectories, we track these time-dependent interaction frame poses. We store them in the pose tuple:

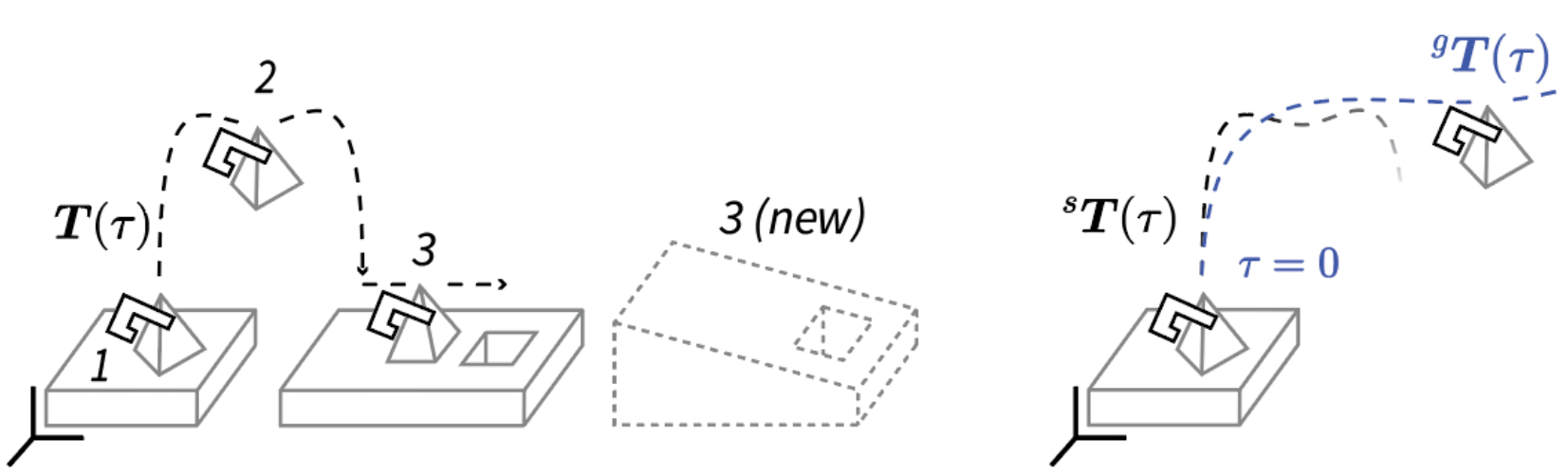

**Fig. 3.** Generalization using constant offsets and trajectory blending. Between the demonstration and the reproduction, segment 1 contains no offset, segment 3 is generalized with a constant pose offset, and the trajectory of segment 2 must be modified to link them. To obtain a generalized pose trajectory ${}^{g}\boldsymbol{T}(\tau)$ for segment 2, we start at the demonstrated pose ${}^{s}\boldsymbol{T}(\tau)$ suited to segment 1 and blend towards the new pose ${}^{e}\boldsymbol{T}(\tau)$ suited to the new segment 3 by interpolating over progress variable $\tau$.

$$ {}_{n}\widetilde{\boldsymbol{P}}^{\chi}(\tau) = \left({}_{n}\widetilde{\boldsymbol{R}}^{\chi}(\tau), {}_{n}\widetilde{\boldsymbol{o}}^{\chi}(\tau)\right). $$

Depending on the frame type of $\{\chi\}$, the variables are expressed with respect to the *ground* or *body*. The demonstrated pose-wrench pair

$$ {}_{n}\boldsymbol{x}^{\chi}(\tau) = \left({}_{n}\widetilde{\boldsymbol{P}}^{\chi}(\tau), {}_{n}\boldsymbol{w}^{\chi}(\tau)\right), $$

will then be used during the reproduction control, expressed in the interaction frame $\{{}_{n}\chi\}$ of each segment.

### B. Geometric Generalization

After modeling a task using demonstration data, the reproduction may contain geometric variations of that task, such as objects with different locations or sizes. The task model may be generalized to the new situation if those variations are known. For example, variations may be estimated before reproduction begins using vision, or during runtime using vision or haptic measurements.

#### 1) Applying Constant Offsets

If all segments of the new situation are offset with respect to the demonstration by the same constant pose, that offset can be directly applied to all segments in the task model.

More commonly, different segments may be offset by different poses (Fig. 3). For example, an object may be picked up from the same location as in the demonstration (no offset) but be inserted into a box in a different location (non-zero offset). In that case, the insertion segments related to the box all have the same non-zero offset.

Offsets may occur in the environment and/or in the grasped object. For example, in a pen-on-board example (Fig. 4), the board may be in a different pose, but the pen may also be grasped differently or have a different size. Depending on in which body the offset occurs (end effector or ground), different components of the interaction frame trajectory must be changed, parameterized by ${}_{n}\boldsymbol{R}^{\chi}$ or ${}_{n}\widetilde{\boldsymbol{R}}^{\chi}(\tau)$ and ${}_{n}\boldsymbol{o}^{\chi}$ or ${}_{n}\widetilde{\boldsymbol{o}}^{\chi}(\tau)$, depending on the frame type (Section IV.A). For the pen-on-board task with frame type $\{\mathcal{GB}\}$, the pen length can be changed with the constant ${}_{n}\boldsymbol{o}^{\chi}$, the orientation of the grasp with ${}_{n}\widetilde{\boldsymbol{R}}^{\chi}(\tau)$, and the orientation of the board with ${}_{n}\boldsymbol{R}^{\chi}$ and ${}_{n}\widetilde{\boldsymbol{o}}^{\chi}(\tau)$.

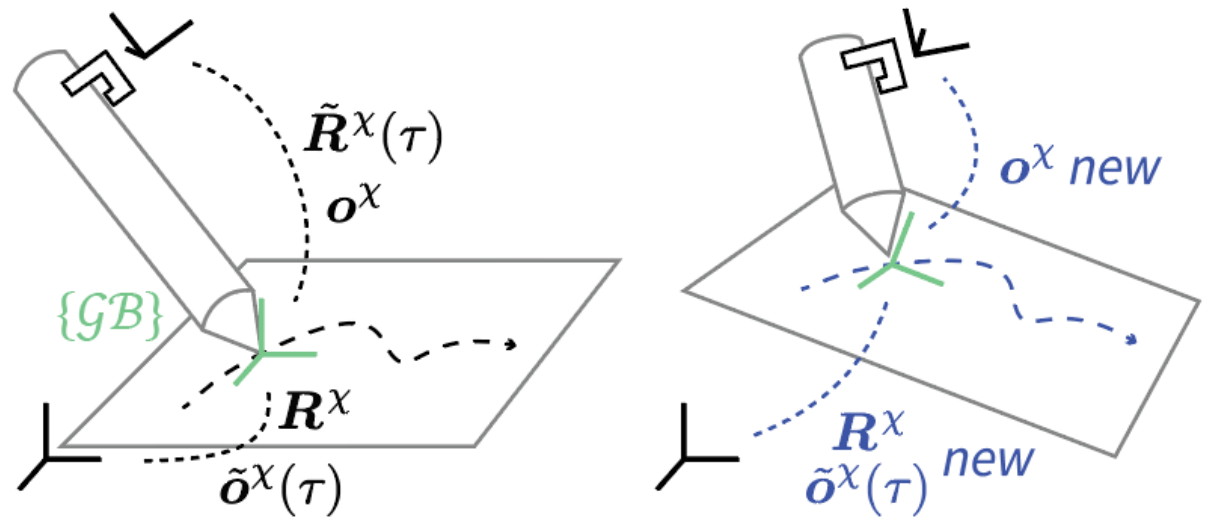

**Fig. 4.** Generalization using constant offsets on interaction frame parameters. Demonstrated trajectories are reproduced by tracking $\widetilde{\boldsymbol{R}}^{\chi}(\tau)$ and $\widetilde{\boldsymbol{o}}^{\chi}(\tau)$ with control expressed in $\{\chi\}$. Tasks may be generalized to a new situation by changing the interaction frame parameters (Table I), using external information obtained beforehand, or during reproduction.

In this work, we assume that only the environment contains geometric variations, while the grasped objects are the same. Therefore, we make adaptations only to the ground-respective components of ${}_{n}\boldsymbol{R}^{\chi}$ or ${}_{n}\widetilde{\boldsymbol{R}}^{\chi}(\tau)$ and ${}_{n}\boldsymbol{o}^{\chi}$ or ${}_{n}\widetilde{\boldsymbol{o}}^{\chi}(\tau)$, depending on the frame type.

#### 2) Blending Trajectories

To link two segments that are offset differently (Fig. 3), the segment in between must be modified. Therefore, we first modify the demonstrated pose trajectory $\boldsymbol{T}(\tau)$ of the body frame with respect to the ground frame. We do so by blending the demonstrated poses transformed to the desired start ${}^{s}\boldsymbol{T}(\tau = 0)$ with the desired trajectory transformed to the desired end ${}^{e}\boldsymbol{T}(\tau = 1)$ over $\tau$:

$$ {}^{g}\boldsymbol{T}(\tau) = {}^{s}\boldsymbol{T}(\tau) \exp\left(\beta(\tau) \ln\left({}^{s}\boldsymbol{T}(\tau)^{-1}\, {}^{e}\boldsymbol{T}(\tau)\right)\right). $$

Here, $\beta(\tau) \in [0,1]$ is a blending function that can be chosen to determine how the trajectory blends from ${}^{s}\boldsymbol{T}(\tau)$ to ${}^{e}\boldsymbol{T}(\tau)$ over $\tau \in [0,1]$. For example, $\beta(\tau) = \tau$ implements linear blending, where the position components are ${}^{s}\boldsymbol{o}(\tau)\,\tau + {}^{e}\boldsymbol{o}(\tau)\,(1-\tau)$. The orientation components are blended analogously.

In this work, instead of linear blending we use a sigmoid function for $\beta(\tau)$, such that ${}^{g}\boldsymbol{T}(\tau)$ behaves like ${}^{s}\boldsymbol{T}(\tau)$ near $\tau = 0$ and like ${}^{e}\boldsymbol{T}(\tau)$ near $\tau = 1$. Thereby, the start and end are largely similar to the original curvatures, such as the directions of motion when making and breaking contact (Fig. 3).

From the newly generalized pose trajectory ${}^{g}\boldsymbol{T}(\tau)$, the new interaction frame trajectory $({}_{n}\boldsymbol{R}^{\chi}, {}_{n}\widetilde{\boldsymbol{R}}^{\chi}(\tau), {}_{n}\boldsymbol{o}^{\chi}, {}_{n}\widetilde{\boldsymbol{o}}^{\chi}(\tau))$ can be obtained for a given frame type. Those interaction frame trajectories are then used during the reproduction control.

## V. Task Reproduction

After modeling a task (Section III) and generalizing it to a new geometric situation (Section IV), the resulting task model can be used to control the robot during reproduction. Various control strategies are viable, but to illustrate the benefits of our proposed task modeling approach we apply established hybrid position/force control (Section V.A) and extend it with contact-aware monitoring and switching (Section V.B and V.C).

### A. Controller Choice

To track the demonstrated poses and wrenches stored in ${}_{n}\boldsymbol{x}^{\chi}(\tau)$, we use a hybrid pose/wrench controller expressed in the identified interaction frames (Fig. 5). We choose pose or wrench control on each axis based on the interaction classes ${}_{n}\boldsymbol{\mathcal{S}} \in \{\mathcal{I}, \mathcal{F}, \mathcal{Z}, \mathcal{C}\}^{12}$ in each segment $n$ of the demonstration.

We track poses for axes that were identified as $\mathcal{F}$ during the demonstration, because those axes contained substantial motion in the twists, but not substantial wrenches. Analogously, we track wrenches for axes that were identified as $\mathcal{C}$. In case the twists and wrenches were both insubstantial ($\mathcal{I}$) or both substantial ($\mathcal{Z}$), choosing the control is less clear. For $\mathcal{I}$, we choose pose control to avoid drift. For quasistatic interactions with static and rigid environments (Section II.A), $\mathcal{Z}$ can only occur due to friction during motion. Similar to $\mathcal{F}$, we assume such motion to be intentional, and therefore also track poses.

For static environments, the classes of the positive and negative axes will be either the same, or one is idle (Section III.B). In case one is idle, we choose control based on the non-idle direction. For example, for ${}^{+}\mathcal{S} = \mathcal{F}$ and ${}^{-}\mathcal{S} = \mathcal{I}$ the control is determined by $\mathcal{F}$ to be pose control.

### B. Hybrid Position/Force Control

The hybrid controller computes a control wrench ${}^{c}\boldsymbol{w}^{\chi} \in \mathbb{R}^{6}$ in the interaction frame $\{\chi\}$, consisting of a pose control term ${}^{p}\boldsymbol{w}^{\chi}$ and wrench control term ${}^{w}\boldsymbol{w}^{\chi}$:

$$ {}^{c}\boldsymbol{w}^{\chi} = \boldsymbol{\lambda} \circ {}^{p}\boldsymbol{w}^{\chi} + (\boldsymbol{1} - \boldsymbol{\lambda}) \circ {}^{w}\boldsymbol{w}^{\chi}\,, $$

which are selected using a binary selection vector $\boldsymbol{\lambda} \in \{0,1\}^{6}$, its complement $(\boldsymbol{1} - \boldsymbol{\lambda})$, and the element-wise product $\circ$.

For axes $i$ that were identified as constrained during the demonstration ($\mathcal{S}_i = \mathcal{C}$), we apply proportional and integral closed-loop wrench control ($\lambda_i = 0$) with feedforward:

$$ {}^{w}\boldsymbol{w}^{\chi} = {}^{w}_{p}\boldsymbol{g} \circ \Delta\boldsymbol{w}^{\chi}(\tau) + {}^{w}_{i}\boldsymbol{g} \circ \int_{o}^{\tau} \Delta\boldsymbol{w}^{\chi}(u)\,\mathrm{d}u + \boldsymbol{w}^{\chi}(\tau), $$

where ${}^{w}_{p}\boldsymbol{g}, {}^{w}_{i}\boldsymbol{g} \in \mathbb{R}^{6}_{+}$ are controller gains, $\Delta\boldsymbol{w}^{\chi}(\tau) = \boldsymbol{w}^{\chi}(\tau) - \widehat{\boldsymbol{w}}^{\chi}$ is the error between the reference wrench from the demonstration $\boldsymbol{w}^{\chi}(\tau)$ and the currently measured wrench $\widehat{\boldsymbol{w}}^{\chi}$.

For axes $j$ that were not identified as constrained during the demonstration ($\mathcal{S}_j \in \{\mathcal{I}, \mathcal{F}, \mathcal{Z}\}$), we apply proportional, integral, and derivative closed-loop pose control ($\lambda_i = 1$):

$$ {}^{p}\boldsymbol{w}^{\chi} = {}^{p}_{p}\boldsymbol{g} \circ \Delta\boldsymbol{p}^{\chi}(\tau) + {}^{p}_{i}\boldsymbol{g} \circ \int_{o}^{\tau} \Delta\boldsymbol{p}^{\chi}(u)\,\mathrm{d}u + {}^{p}_{d}\boldsymbol{g} \circ \frac{d\Delta\boldsymbol{p}^{\chi}(\tau)}{d\tau}, $$

where the pose error expressed in the interaction frame (Fig. 5) is $\Delta\boldsymbol{p}^{\chi}(\tau) = \widehat{\widetilde{\boldsymbol{R}}}^{\chi} \circ \left[\Delta\boldsymbol{\theta}^{\chi}(\tau), \widetilde{\boldsymbol{o}}^{\chi}(\tau) - \widehat{\widetilde{\boldsymbol{o}}}^{\chi}\right] \in \mathbb{R}^{6}$. The rotation vector error is $\Delta\boldsymbol{\theta}^{\chi}(\tau) = \left[\ln\left(\Delta\boldsymbol{R}^{\chi}(\tau)\right)\right]^{\vee}$, using the rotation matrix error $\Delta\boldsymbol{R}^{\chi}(\tau) = \left(\widehat{\widetilde{\boldsymbol{R}}}^{\chi}\right)^{T} \widetilde{\boldsymbol{R}}^{\chi}(\tau)$.

### C. Monitoring and Switching Control

Section III.C derived the transition conditions between segments ${}_{n}\boldsymbol{\pi} = \{{}^{\wedge}c, {}^{\vee}c, {}^{>}\tau\}^{12}$, for positive and negative rotations about and translations along the three axis of an interaction frame. The conditions are making contact (${}^{\wedge}c$), breaking contact (${}^{\vee}c$), and tracking completion (${}^{>}\tau$). To reproduce tasks in a contact-aware manner, we define several general-purpose conditions that must be fulfilled before the control switches to the next segment (Fig. 5).

#### 1) Conditions for Tracking Completion

If the transitions conditions between two segments only contain tracking progress completion ${}_{n}\boldsymbol{\pi} = \{{}^{>}\tau\}^{12}$, the preceding segment may be considered purely a pose/wrench tracking segment, since there are no contact changes (Fig. 5a). The first condition to consider that segment completed is when the reference progress is at its end:

$$ \tau = 1. \tag{2} $$

Furthermore, there may be tracking requirements depending on the specific task. For example, grasping a small object may require relatively small pose errors before starting the grasp. Therefore, the second requirement is that pose tracking errors $\Delta\boldsymbol{p}^{\chi}(\tau)$ with respect to the end of the segment ($\tau = 1$) must be within prespecified tolerances

$$ \left|\Delta p^{\chi}_{j}(1)\right| < {}^{*}p_{j}, \tag{3} $$

where ${}^{*}\boldsymbol{p} \in \mathbb{R}^{6}_{+}$ are prespecified tolerances on the pose errors of the pose-tracked axes $j$. These tolerances may be chosen for a large range of tasks. On wrench-tracked axes $i$, we do not check the pose errors since we assume contact to be more important.

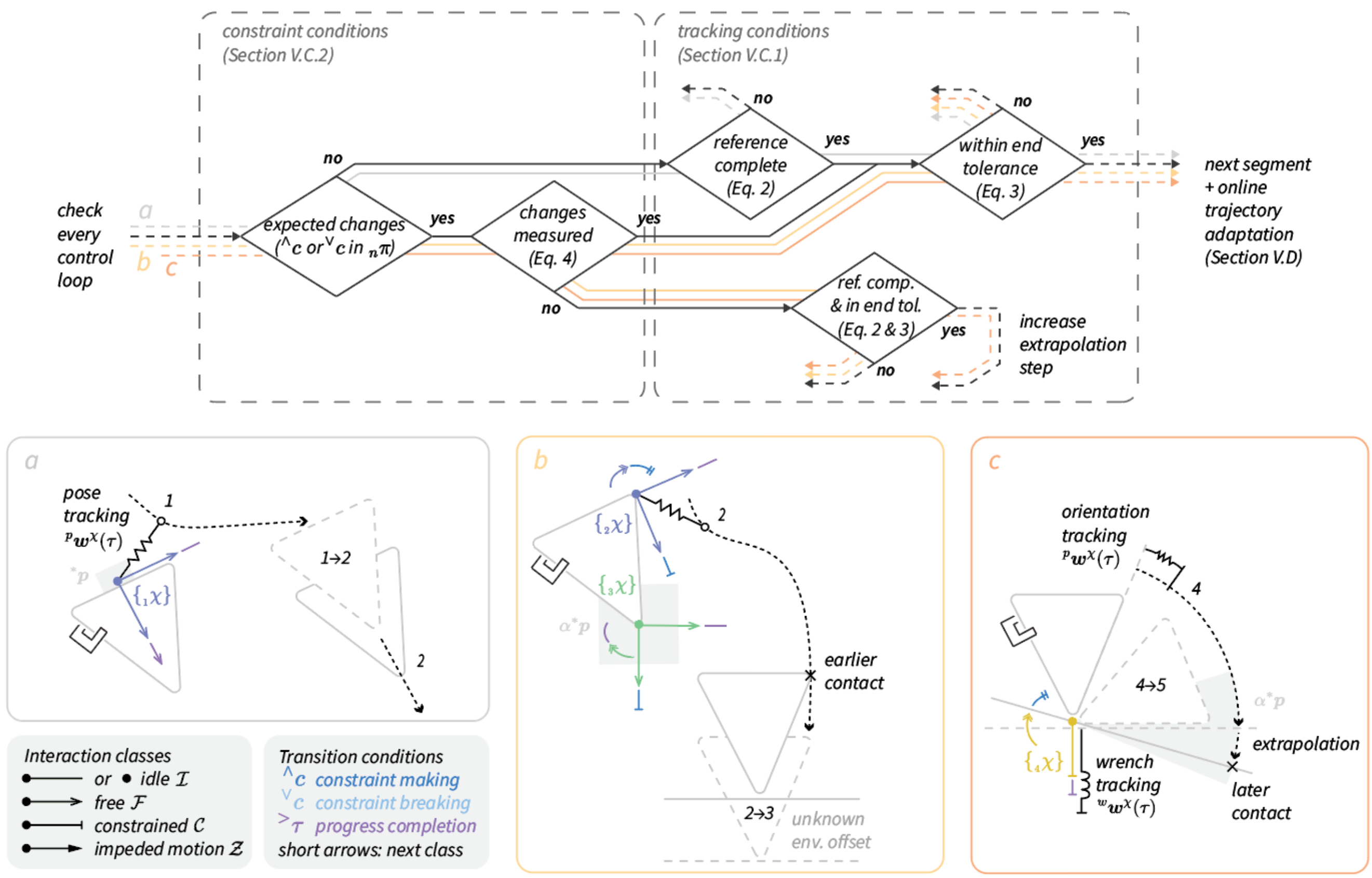

**Fig. 5.** Reproduction logic flowchart and examples of a robot manipulating a triangle to contact and align with a plane. The lines through the flowchart represent the logic that is triggered during the three example segments. (a) Tracking in free space using only pose control, no contact is expected. The control is shown with springs and expressed in the identified interaction frames $\{_n\chi\}$. Tracking finishes when both the progress variable completes (${}^{>}\tau$, (2)) and the endpoint is within the pose tolerance (${}^{*}\boldsymbol{p}$, (3)). (b) Transitioning from free space to contact, where transition conditions (${}^{\wedge}c$) can be more compactly expressed in $\{_3\chi\}$ than in $\{_2\chi\}$ (Section III.C). Because a contact change is expected, the tolerance is relaxed ($\alpha^{*}\boldsymbol{p}$), providing robustness to earlier-than-expected contact. (c) A force controller maintains the contact, and a pose controller maintains the horizontal position of the contact point and rotates the triangle. Similar to (b), the tolerances are relaxed. The orientation is extrapolated to explore for the rotation contact, providing robustness to later-than-expected contact.

*2) Conditions for Contact-Aware Transitions*

In addition to tracking progress completion (${}^{>}\tau$), the transition conditions $_n\boldsymbol{\pi} = \{{}^{\wedge}c, {}^{\vee}c, {}^{>}\tau\}$[12] may contain the making (${}^{\wedge}c$) and/or breaking of constraints (${}^{\vee}c$). In such cases, we also require that those expected constraint changes are measured online. We do so by monitoring the interaction classes $\hat{\mathcal{S}}_h$ online, where $h$ are the axes along which making or breaking of constraints is expected to occur:

$$\hat{\mathcal{S}}_h = \mathcal{C} \quad \text{if} \quad \pi_h = {}^{\wedge}c, \qquad \text{or} \qquad \hat{\mathcal{S}}_h \in \{\mathcal{F}, \mathcal{Z}\} \quad \text{if} \quad \pi_h = {}^{\vee}c. \tag{4}$$

In those cases, we allow those constraint transitions to be the guiding factor in segment completion by relaxing all completion tolerances on the right-hand side of (3) by a multiplier $\alpha > 1$. Thereby, providing robustness to unknown variations in where contact changes occur (Fig. 5b-c).

*a) Earlier-Than-Expected Constraint Transitions*

The expected contact transitions (4) may occur ahead of the transitions in the reference demonstration ($\tau < 1$, Fig. 5b). For example, when approaching a contact plane, and the plane is closer than demonstrated. In such cases, we switch control to the next segment regardless of $\tau$, if the errors are within the relaxed tolerances (3).

*b) Later-Than-Expected Constraint Transitions after Exploration*

Alternatively, the expected contact transitions (4) may not have occurred even though the end of the reference trajectory has been reached (2), and the tracking error is within the end tolerance (3). In such cases, exploration is needed to find the expected transitions. Here, we implement straightforward exploration strategies of poses and wrenches, to cover common transitions.

For pose exploration, we linearly extrapolate the last one-fifth of the reference pose $\tilde{\boldsymbol{P}}^{\chi}(\tau)$, by extrapolating the position vector $\tilde{\boldsymbol{o}}^{\chi}(\tau)$ and rotation vector $\tilde{\boldsymbol{\theta}}^{\chi}(\tau)$ at the same time. For example, when a contact plane is further than demonstrated, the exploration continues the approach with a similar velocity and direction until contact is encountered (Fig. 5c).

For wrench exploration, we increase the reference wrench $\boldsymbol{w}^{\chi}(\tau)$ over time, but only on axes $h$ along which a breaking constraint ($\pi_h = {}^{\vee}c$) is expected. For example, if an object cannot be dislodged using the reference wrench from the demonstration, the wrench is increased in the direction of the desired dislodging.

Sufficient pose (and wrench) exploration may then cause the expected making or breaking of constraints to occur. The exploration continues only as far as the relaxed tolerances allow (3).

### *D. Online Trajectory Adaptation using Haptic Information*

A segment may be completed in a different pose than during the demonstration due to previously unknown variations in the geometry (Fig. 5b-c). In such cases, it may be necessary to adapt the next segment accordingly using the currently measured pose offset. To do so, we generalize the task model online during segment transitions (Section IV.B).

Whether to generalize the next segment using a constant offset (Section IV.B.1) or trajectory blending (Section IV.B.2) depends on whether the currently measured offset also applies to the complete next segment. If the current segment and next segment both contain a constraint, we assume both segments are in contact with the same locally offset geometry of the environment and apply the constant offset (Fig. 3 segment 3). Alternatively, if the next segment does not contain a constraint, we assume it is in free space and blend the trajectories back to the original demonstration, assuming the next segments' geometric features are unrelated to the currently offset segment.

## VI. Evaluation

We evaluated our LfD method on several complex, sequential, and contact-rich robot tasks (Figures 7-9 and video[3]). A human operator demonstrated each task only once by guiding the robot in low impedance mode. After autonomous task modeling, the tasks were reproduced with geometric variations such as objects poses.

### *A. Reproduction Evaluation*

To illustrate how task modeling (Section III) supports LfD, we started with a basic controller and implement our contact-aware controller and monitor step-by-step (Section V). We used five steps, where each step uses more information than the previous one. We reproduced each task variation 10 times and reported the success rates to increasingly large geometric variations. If one of the controller steps failed one of the reproductions, we enabled the next controller step and re-attempted the full task.

The first three controller steps were used to evaluate *robustness to unknown* geometric variations. The first controller step was pure pose control expressed in the end effector frame (Section V.B) with tracking monitoring only (Section V.C.1). The second step was hybrid control expressed in the interaction frames (Section V.B) also with tracking monitoring only (Section V.C.1). These two steps use the top path of Fig. 5. The third step adds contact-aware monitoring (Section V.C.2) to the second step, using all paths of Fig. 5.

When the robustness of these three controller steps was no longer sufficient for the increased and unknown geometric variations, we evaluated *adaptivity using information obtained online* using one additional controller step. The fourth step added online trajectory adaptation (Section V.D) to the preceding contact-aware hybrid controller. Upon switching the control to the next segment, the trajectories are generalized to the new situation using the currently measured poses. This fourth controller uses all conditions of the reproduction logic (Fig. 5).

When the adaptivity using information obtained online was no longer sufficient for the increased geometric variations, we assumed them to be *known a priori*, such that task models could be *generalized offline*. We assumed those variations were known accurately through external information, such as vision. In the fifth step, we first generalized the task model (Section IV.B) and then used the hybrid controller. Because the variations were known beforehand, steps three and four were not needed anymore.

### *B. Implementation*

We used a Franka Research 3 robot with a Franka Hand gripper (Franka Robotics, Munich, Germany), a SensONE force torque sensor (Bota Systems, Zurich, Switzerland) between robot and gripper, and household objects to interact with. We used a custom end effector in the last task (Fig. 8).

Our task modeling[1] and monitor/control[2] methods were implemented in Python 3.11.4 and are publicly available. The monitor/control (Section V) ran at 100 Hz and interfaced with the robot over ROS2 Humble Hawksbill on an Ubuntu 22.04 computer.

We compensated for the gravitational effect of the gripper in the wrench sensor measurements by subtracting the gripper mass at its center of gravity. The gripper was operated by keyboard presses, which resulted in a new segment that lasted until the opening or closing was complete. During the reproduction of those grasping segments, the robot was fully pose-controlled.

The parameters used during the experiments are shown in Table II. The task modeling thresholds were taken from our previous work [20]. The transition tolerances (3) were chosen such that reproductions in the same geometric situations as the ones demonstrated resulted in 100% success rates. Thereby, the reproduction tracking was accurate enough such that segments that are sensitive to pose errors, such as peg insertion or grasping smaller objects, did not limit reproduction success.

[1] https://github.com/ET-BE/ReFrameId
[2] https://github.com/ET-BE/ReFrameCtrl
[3] https://utwente.yuja.com/v/interaction_lfd

We evaluated three main tasks: cabinet opening, bolt screwing, and surface following. We split these tasks into subtasks with sustained contact, separated by free space reaching and grasping segments. Because the free space segments did not limit reproduction success, the subtasks may be evaluated individually. We do not discuss segments in which the robot is static, such as while waiting for the gripper to start closing, because they do not limit reproduction success.

### *C. Cabinet Opening*

This task consisted of unlocking a hinge latch, unlocking a barrel latch, and opening a door (Fig. 6).

#### *1) Hinge Latch Opening*

This subtask (Fig. 6a) consisted of approaching the handle (**1**), aligning the gripper (**3**), closing the gripper (**5**), rotating it (**6**) until reaching a constraint at the fully open position (making a rotation constraint $^\wedge c$, **6**→**7**), opening the gripper (**9**), and retreating the gripper (**10**). During the pure rotation segment (**6**), the interaction frame was identified in the rotation axis.

The pose controller reproduction was successful for angle offsets up to 5.6 deg. For a 8.4 deg offset, the mechanism forced deviation from the demonstration reference trajectory at the open position, sometimes outside of the allowed tolerances, failing that transition (**6**→**7**) in three cases. Whether the tolerance was met at the open position depends on the tolerance at the start of the rotation segment (**6**) and thus depends on the accuracy during the previous grasping (**5**). In addition, the stiff pose controller caused substantial wrenches, which in two cases led to the gripper slipping off the handle in the open segments (**6**, **7**, **8**). Thereby, resulting in a 50% success rate.

The hybrid controller's wrench control was more compliant and avoided the gripper slipping off the handle in the opened hinge segments, leading to an increased 70% success in those segments (**6**, **7**, **8**).

After enabling online trajectory adaptation upon segment switching, the subsequent trajectories were adapted to the new open position of the door, and the task was completed consistently. The offsets may be increased to 17.0 deg, but for larger offsets the handle could not be grasped anymore because the gripper collided with the door (**5**).

For such larger $17.0 - 30.0$ deg offsets, offline generalization was needed to modify the trajectory a priori to avoid collision by approaching the handle perpendicular to the cabinet.

#### *2) Barrel Latch Opening*

This subtask (Fig. 6b) consisted of approaching a handle (**1**), closing the gripper (**3**), arching the handle down (**5**), sliding the bolt open (**7**) until a contact was made (making a linear constraint $^\wedge c$, **7**→**8**), arching the handle upward (**9**) until a contact was made (making a linear constraint $^\wedge c$, **9**→**10**), opening the gripper (**12**), and retreating the gripper (**13**).

While moving the latch, the handle knob could rotate slightly in the gripper fingers. Therefore, the end effector was not rigidly fixed to the handle knob, and our assumption on rigidly grasped objects does not hold (Section II.A). However, in quasi-static conditions all demonstrated forces must pass through the bolt handle, and an interaction frame at either end of the handle, i.e. in the knob or in the bolt axis, minimizes the moments in that frame. Therefore, the interaction frames were still identified in the bolt axis handle (**5**, **7**, **9**). Contact was made when sliding the barrel latch horizontally to the open position (**7**→**8**) and fully upward against the cabinet (**9**→**10**).

The pose controller reproduction was successful for angle offsets up to 2.7 deg. For a 5.6 deg latch offset, the latch did not slide to the fully open position because the expected and real bolt axes were misaligned. Therefore, the handle was blocked from arching upward (**9**), thereby getting stuck. The same happened when using the hybrid controller.

After enabling contact-aware switching, the extrapolation caused momentary contact at the open position (**7**→**8**). However, because the subsequent trajectories (**8**, **9**) were not adapted to the new situations, the upward arch of the handle (**9**) was reproduced as demonstrated and also blocked, similar to the pose controller.

After enabling online trajectory adaptation upon segment switching, the next segments' trajectories (**8**, **9**, **10**, **11**) were generalized to the new fully open position, and the tasks were completed. For an increased 11.3 deg offset the same controller was still successful, but for a 14.1 deg offset the gripper could not reach the pose to grasp the handle knob anymore (**3**) due to collision with the door.

Offline generalization was needed for such larger $14.1 - 30.0$ deg offsets, similar to 1).

#### *3) Door Opening*

This subtask (Fig. 6c) consisted of approaching the door handle (**0**), closing the gripper (**2**), swinging the door open (**4**) until a contact was reached (making a rotation constraint $^\wedge c$, **4**→**5**), and opening the gripper (**7**). During the door swing (**4**), the interaction frame was identified in the rotation axis of the door.

The pose controller reproduction was successful for cabinet offsets of up to 1.4 deg, equivalent to a 10.0 mm offset of the rotation axis in the door hinge. This resulted in an approximately 10.0 mm tracking error at the end of the opened arc, larger than the required 4.0 mm tracking tolerance, but because the handle slipped slightly in the gripper the controller could reduce the tracking error below the required tolerance and finish the task. For a 2.8 deg cabinet offset (20.0 mm axis offset), the end effector in the open position was constrained into a pose that is not within the allowed tolerances of transition (**4**→**5**). Due to the stiff pose control, the door was also subjected to substantial wrenches, and the gripper was close to slipping off the door handle.

After enabling compliant hybrid control and contact-aware switching, the relaxed tolerances allowed the failing transitions (**4**→**5**) to be successful. For a 4.2 deg cabinet offset (20.0 mm axis offset), the controller did not reach the fully open door (**6**).

Offline generalization was needed for such larger $4.2 - 30.0$ deg offsets, similar to 1) and 2).

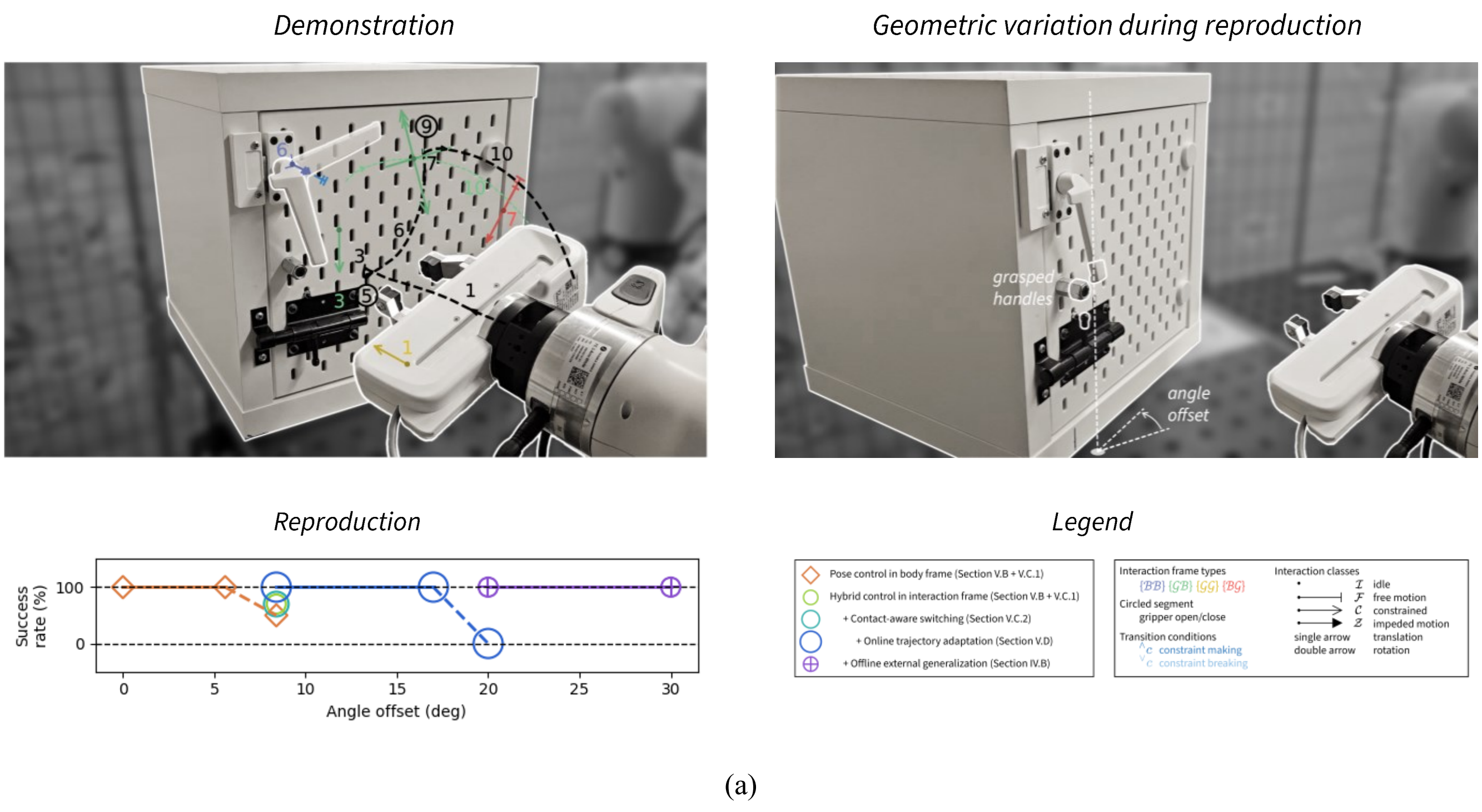

**Barrel latch opening**

Success rate (%)

Angle offset (deg)

(b)

**Door opening**

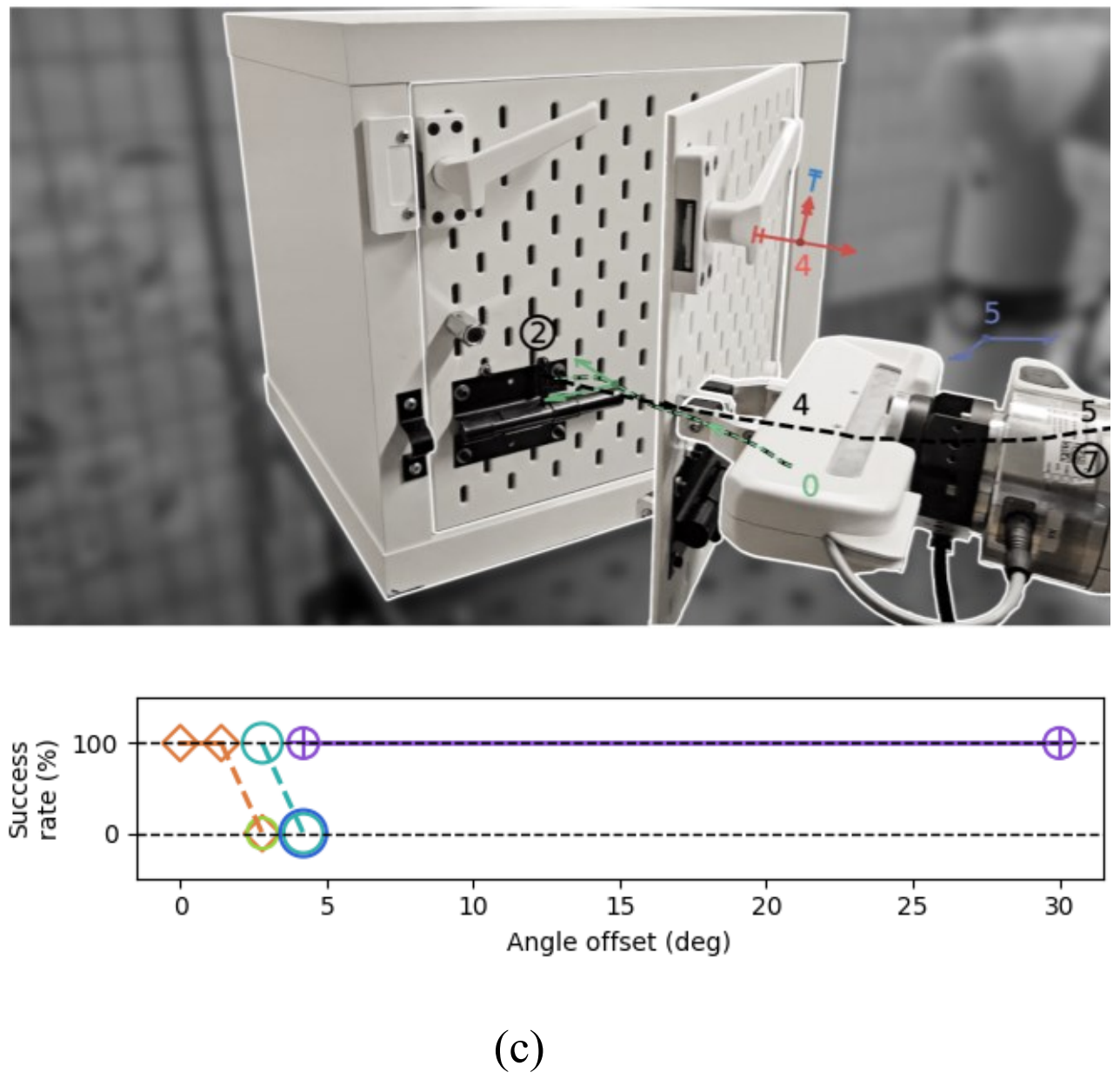

Success rate (%)

Angle offset (deg)

(c)

**Fig. 6.** Opening a cabinet by (a) unlocking a hinge latch, (b) unlocking a barrel latch, and (c) opening a door. The end effector frame is located between the gripper and wrench sensor, and its positions are shown in black. Interaction frames and their paths through space are colored. Numbers indicate segments, and idle waiting segments are not shown. The success rates of different controllers are evaluated at increasingly large geometric variations by rotating the entire cabinet (a, right). The cabinet is rotated around an axis such that the grasping handles of the hinge latch, barrel latch, and door stay in approximately the same position.

**Bolt picking**

**Hole search and screwing**

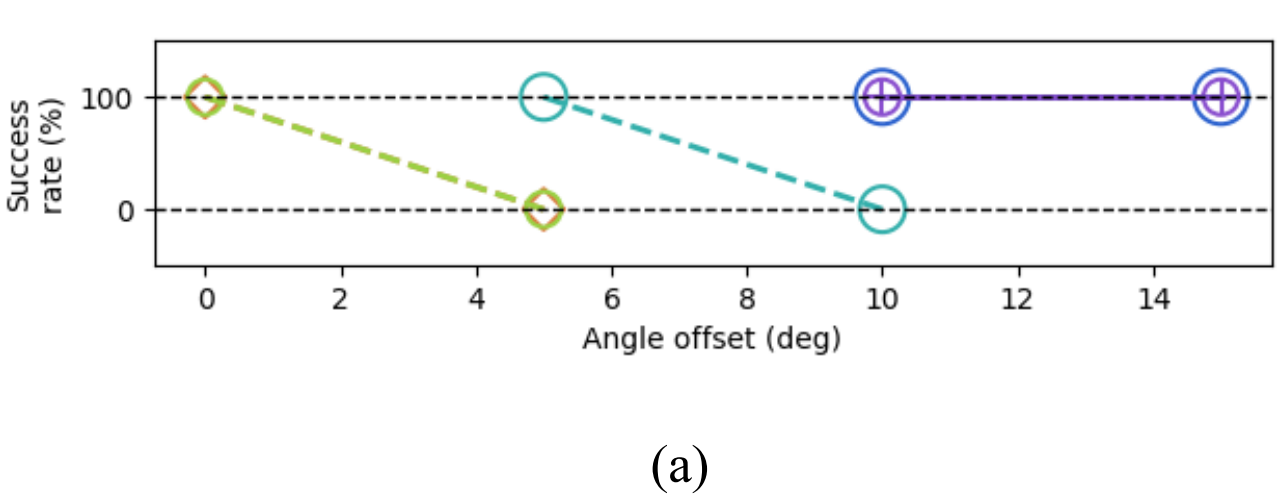

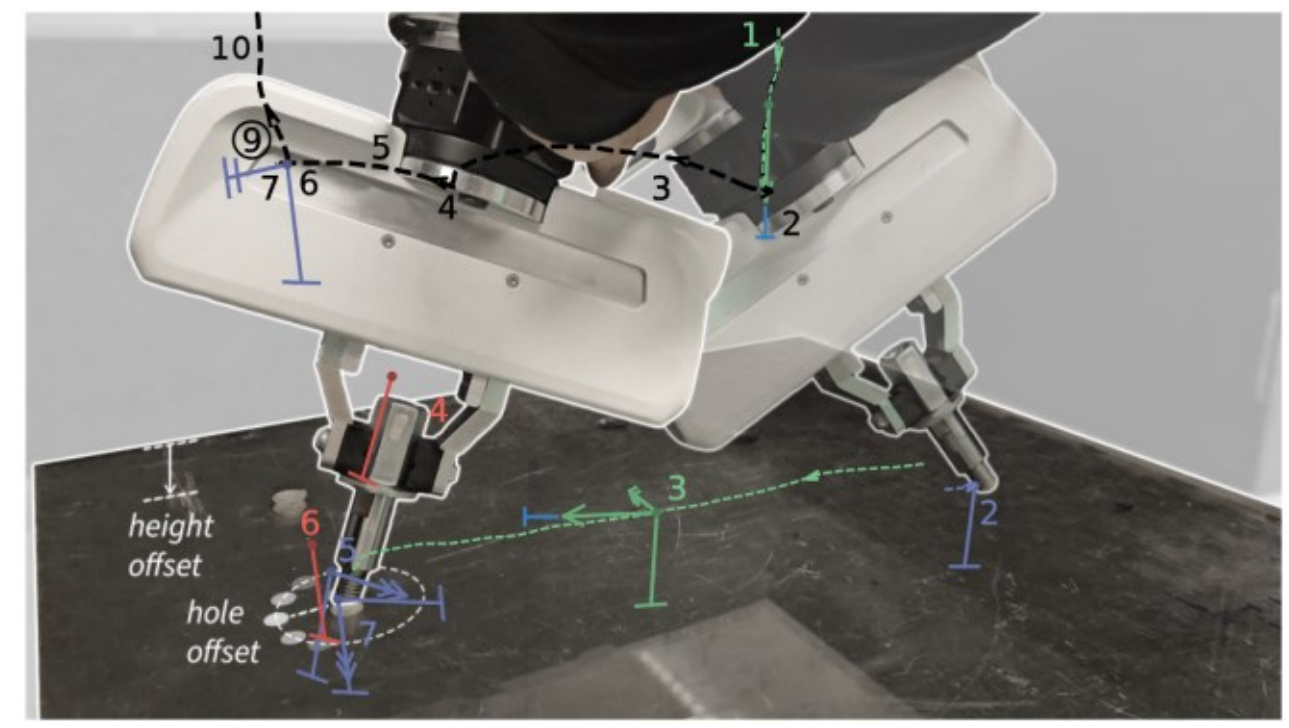

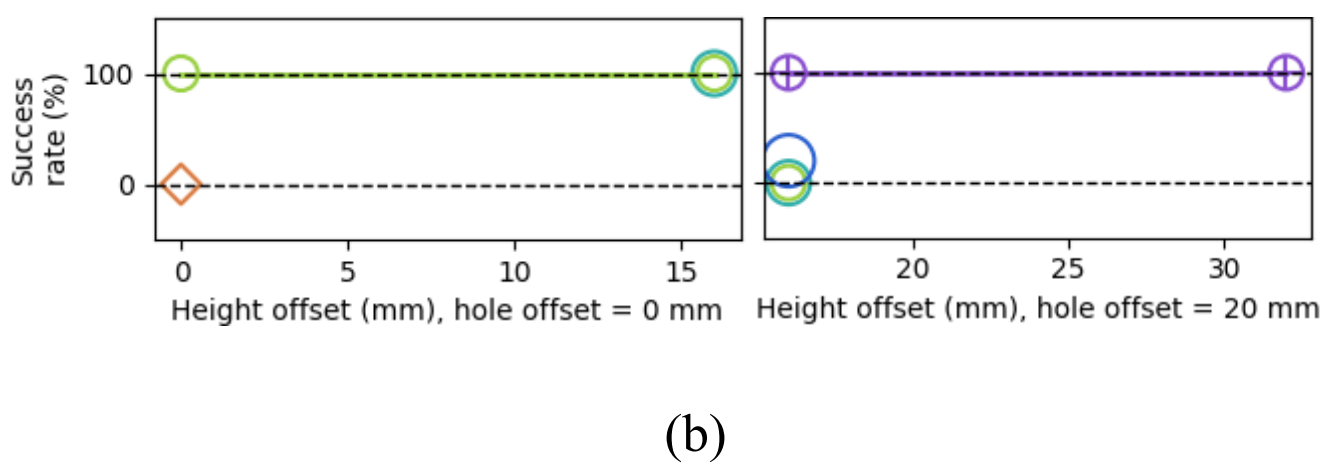

(a)

(b)

**Fig. 7.** Picking a bolt (a) and hole searching, aligning, and screwing the bolt into a threaded hole in the plate (b). The success rates of different controllers are evaluated at increasingly large geometric variations by (a) slanting the surface that holds the bolt, (b) lowering the plane that contains the hole, and replacing the hole with another one on a 20 mm circle around the demonstrated one. See Fig. 6a for the legends.

## D. *Bolt Screwing*

This task consisted of picking a bolt and screwing it into a threaded hole (Fig. 7).

### 1) *Bolt Picking*

This subtask (Fig. 7a) consisted of approaching a plane (**1**) until contact was made with one gripper finger (making a linear constraint $^{\wedge}c$, **1→2**), rotating around that contact (**3**) until the second finger contacts the plane (making a rotation constraint $^{\wedge}c$, **3→4**), closing the gripper (**5**), and lifting up the bolt (**6**). While the first finger was in a point contact with the plane (**2**, **3**), the interaction frame was identified in the contact point.

The pose controller reproduction was successful if there were no to minimal geometric variations. Introducing a 5 deg plane offset causes the rotation contact (**3→4**) to occur earlier than expected when compared to the reference trajectories. Completing the reference trajectories would require mechanical interference, and so transition (**3→4**) could not be completed because the required tracking tolerances could not be met.

The hybrid controller succeeded on the translation tolerances but still failed on the rotation tolerances of the same transition (**3→4**), because the controller was expressed in the contact point about which rotation occurs.

After enabling contact-aware transitions, the relaxed tolerances allowed the task to be completed. For an increased 10 deg plane offset, the reproduction failed because meeting the transition tolerances from the constrained grasped bolt to free space (**5→6**) would require mechanical interference.

After enabling online trajectory adaptation, the trajectories were adapted based on the online pose measured at the 10 deg offset at the failing transition (**5→6**). Thereby, avoiding the need for mechanical interference during tracking and allowing the task to be completed. For such larger $10-15$ deg offsets, adapting the trajectories online or offline both resulted in success.

### 2) *Hole Search and Screwing*

This subtask (Fig. 7b) consisted of approaching a plane (**1**) until the bolt makes contact (making a linear constraint $^{\wedge}c$, **1→2**), sliding the bolt over the plane (**3**) until it catches on the hole (making a linear constraint $^{\wedge}c$, **3→4**), aligning the bolt with the hole perpendicular to the plane (**5**), screwing it in with a half turn (**7**), opening the gripper (**9**), and retreating the gripper (**10**). During the contact segments, the interaction frames were identified in the contact point between bolt and plane.

The pose controller reproduction was unsuccessful even without geometric variations, because it did not maintain the contact necessary to catch on the hole and screw the bolt into it.

The hybrid controller was successful without geometric variations, because the force controller maintained contact. After lowering the plane height by 16 mm the pose controllers used to approach the plane (**1**) halt above the plane where contact occurred in the demonstration (**1→2**). After the end point tolerances were met in free space, the wrench controller perpendicular to the plane was enabled and contact was made for the remainder of the task until the gripper opened again (**9**). Because wrench controllers are robust to variations in the plane height, the tasks were completed successfully.

***Dislodging by pulling***

***Surface contouring***

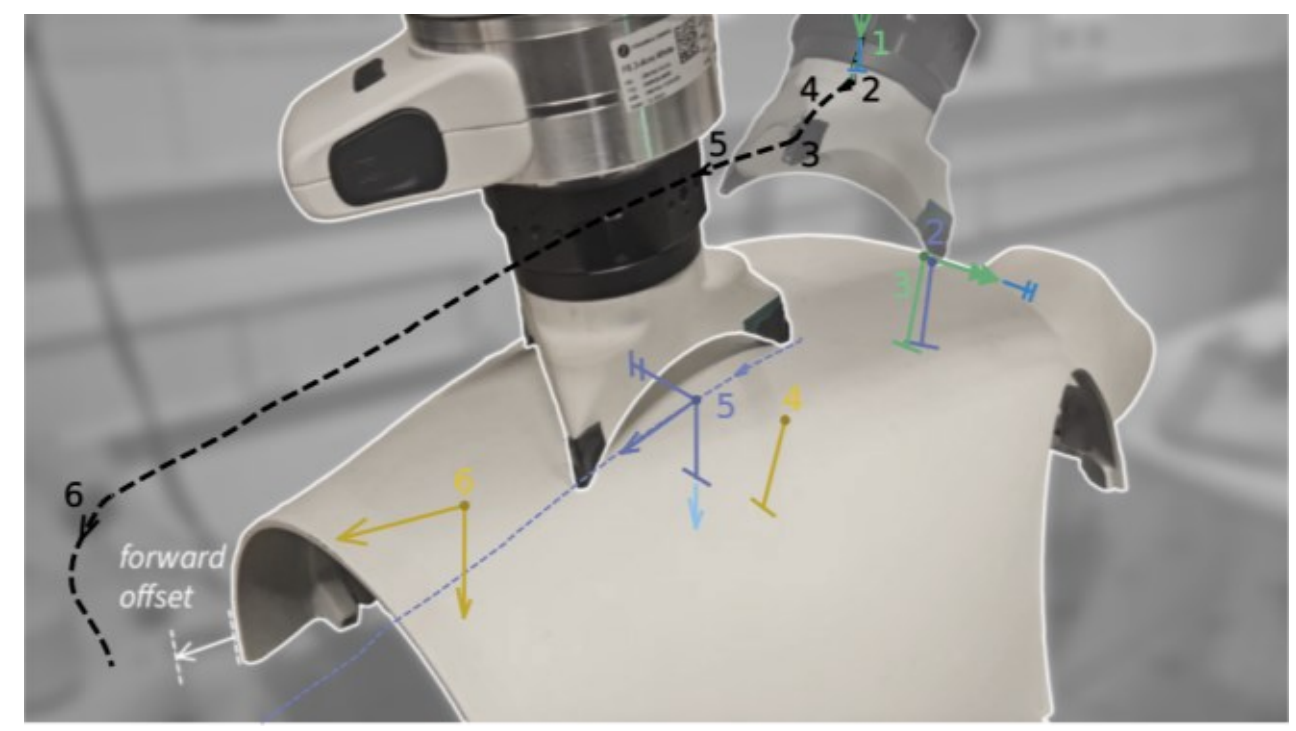

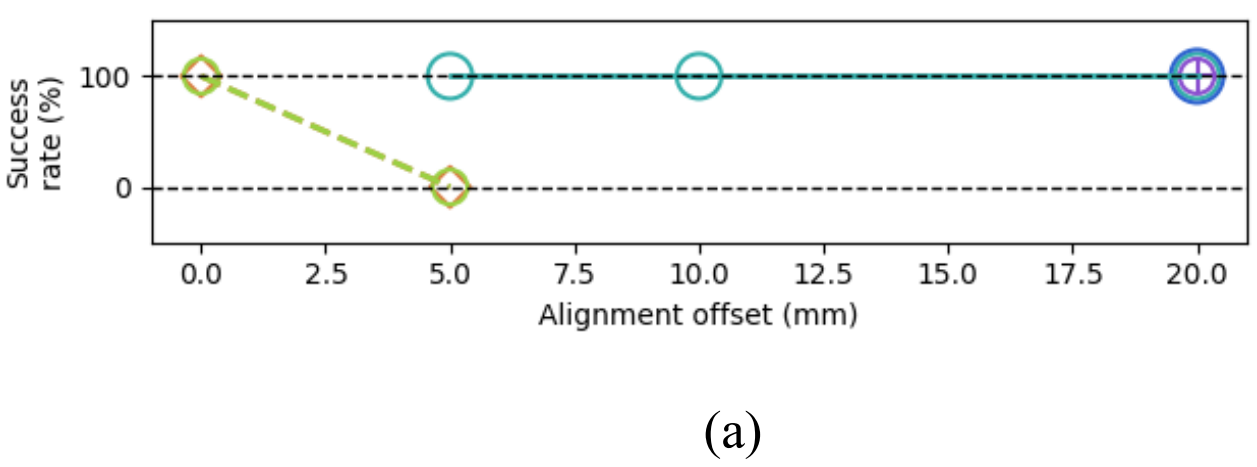

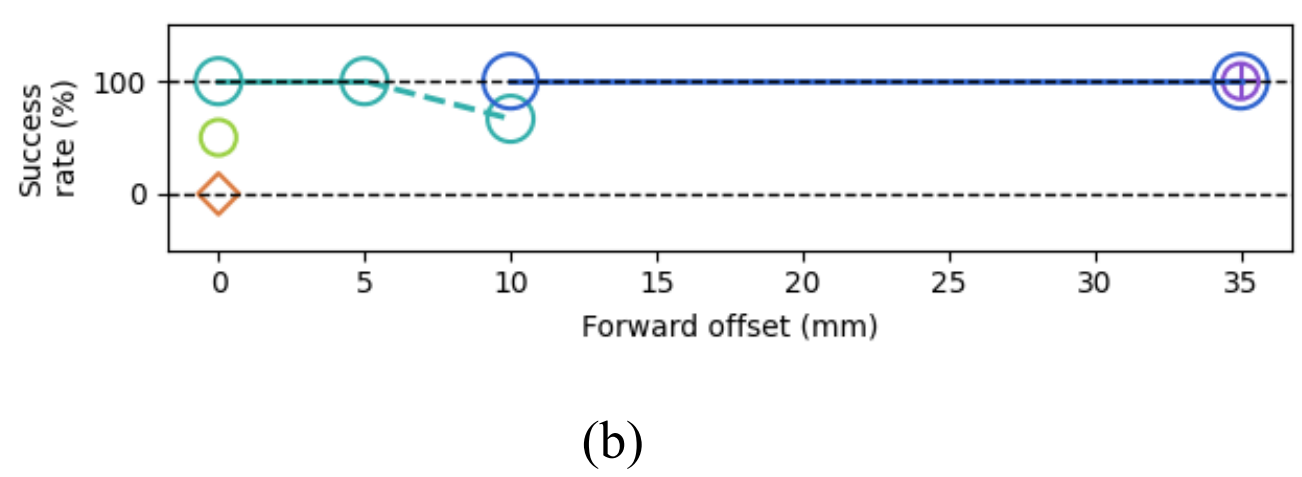

(a) (b)

**Fig. 8.** Dislodging the three-pronged end effector from being clamped between the metal bar and a white block with rubber-rubber contact (a) and using it to follow a three dimensional surface (b). The success rates of different controllers are evaluated at increasingly large geometric variations by (a) offsetting the clamp, thereby increasing maximum stiction force and (b) moving the surface. See Fig. 6a for the legends.

After enabling contact-aware transitions, the approach trajectory (**1**) was extrapolated until contact with the plane was made (**1→2**), instead of relying on the wrench controller for contact. With the hole at the original position, the tasks were also completed successfully. When the hole was offset on a circle with a 20 mm radius, two out of the ten holes (**3→4**) in line with the direction of approach could be found using linear extrapolation. However, because the segments after the hole contact (>**4**) expected the hole in the original position, that tolerance was not met (**4→5**) and the transitions failed.

After enabling online trajectory adaptation upon segment switching, those segments (>**4**) were adjusted accordingly such that the two out of ten hole tasks could be completed, resulting in a 20% success rate. To find the other holes, more elaborate exploration than linear extrapolation is required.

Alternatively, if all ten hole positions were known beforehand, e.g. from a vision system, offline generalization could be used to adapt the trajectories beforehand to find all holes, resulting in a 100% success rate.

### *E. Surface Following*

This task consisted of dislodging a clamped three-pronged end effector by pulling, and then contouring its prongs over a three-dimensional surface until contact was lost off the edge (Fig. 8). We used a custom end effector, different than the robot gripper. During surface following, the end effector maintained contact at its three prongs, and orientations and translations were coupled through contact.

#### *1) Dislodging by Pulling*

This subtask (Fig. 8a) consisted of pulling on a constrained end effector (**1**) hard enough until it was dislodged (losing a linear constraint $^{\vee}c$, **1→2**), moving while contacting the environment (**2**), and lifting into free space (**3**). Lifting out of the environment contact was not identified as losing constraints $^{\vee}c$, but as tracking completion $^{>}\tau$, because the constraints and the motion are in opposite directions.

If the clamp was offset perpendicular to the dislodging direction (Fig. 8a), the pose control caused additional forces in that direction. With the end-effector pushing against the clamp surface, the maximum static friction increases, and an increased force is required for dislodging.

The pose and hybrid controller reproductions were successful without any offsets in the clamp position. For a 5 mm offset, they both successfully dislodged the end effector, but could not switch to the next segment (**1→2**) because the required tolerances were not met.

After enabling contact-aware switching, the task could be completed. For 10 mm to 20 mm offsets, the pulling wrenches were explored automatically (increased in the $^{\vee}c$ direction) to overcome the increased maximum static friction.

Alternatively, by enabling online trajectory adaptation, if the offset was measured online (or offline generalization if it was known beforehand) the task model could be adjusted to pull from the measured offset position, resulting in success without needing wrench exploration.

TABLE II

PARAMETERS USED DURING THE EXPERIMENTS

| | Parameter | Symbol | Orientation | Displacement | Moment | Force |
|---|---|---|---|---|---|---|
| Modeling | Thresholds | ${}^{*}\boldsymbol{s}$ | $200\ \mathrm{mrad\ s^{-1}}$ | $20\ \mathrm{mm\ s^{-1}}$ | $0.3\ \mathrm{N\ m}$ | $5\ \mathrm{N}$ |
| Reproduction | Tolerances | ${}^{*}\boldsymbol{p}$ | $30\ \mathrm{mrad}$ | $4\ \mathrm{mm}$ | - | - |
| | Gains proportional | ${}_{p}\boldsymbol{g}$ | $25\ \mathrm{N\ m\ rad^{-1}}$ | $1200\ \mathrm{N\ m^{-1}}$ | $0.2$ | $0.2$ |
| | Gains integral | ${}_{i}\boldsymbol{g}$ | $8\ \mathrm{N\ m\ (rad\ s)^{-1}}$ | $400\ \mathrm{N\ (m\ s)^{-1}}$ | $5.0\ \mathrm{s^{-1}}$ | $0.2\ \mathrm{s^{-1}}$ |
| | Gains derivative | ${}_{d}\boldsymbol{g}$ | $0.8\ \mathrm{N\ m\ s\ rad^{-1}}$ | $40\ \mathrm{N\ s\ m^{-1}}$ | - | - |

dead time $= 0.3$ s, relaxed tolerance multiplier $\alpha = 20$, pose extrapolation ratio $= 0.2$, wrench exploration ratio $= 0.1$

*2) Surface Contouring*

This subtask (Fig. 8b) consisted of approaching a surface (**1**) until contact was made with one of the three prongs (making a linear constraint ${}^{\wedge}c$, **1→2**), rotating about the prong (**3**) until all three contacted the surface (making a rotation constraint ${}^{\wedge}c$, **3→4**), and following the surface (**5**) until the prongs fell off the edge (losing a linear constraint ${}^{\vee}c$, **5→6**). When in contact, the interaction frames were identified near the prong contacts.

The pose controller reproduction did not consistently maintain contact at the three contact points even without any geometric offsets, and may therefore be considered unsuccessful for e.g. a polishing task.

The hybrid controller successfully maintained contact during surface contouring. However, falling off the edge (**5→6**) did not always occur within the required tolerances on the surface perpendicular to the direction of motion due to tracking errors, resulting in a 50% success rate.

After enabling contact-aware switching, the lowered tolerances resulted in success up to a 5 mm offset of the surface in the direction of the contouring motion. For 10 mm offset, the task was completed consistently but one of the three prongs lost contact several times during contouring, and so we consider the task 66% successful. After enabling online trajectory adaptation all prongs maintain contact, because the trajectories are adapted to the new starting pose when contacting the contour.

## VII. DISCUSSION

This work illustrates how explicitly modeling physical interactions benefits LfD of robotic manipulation. We extended our previous work on identifying *what* physical interaction occur *where* and *when* [20], with contact-aware transitions (Section III), trajectory generalization (Section IV), and contact-aware monitoring and control during task reproduction (Section V).

### A. Explicitly Modeling Physical Interactions for LfD

We illustrated that modeling physical interactions allows for LfD with explicitly implemented robustness, generalization, and adaptivity using only a single demonstration. These features can be explicitly implemented because our approach is human-interpretable, in contrast to approaches in the literature that rely on stochastic and/or intractable methods (e.g. black box approaches). Furthermore, quantitative evaluations of robustness, generalization, or adaptivity to increasingly large variations are often lacking. Therefore, such approaches may result in fewer constructive insights, such as the reasons for failing or inconsistent reproduction. In contrast, our method is interpretable and highly consistent in quantitative evaluations, mostly resulting in either 0% or 100% success rates. Therefore, our work provides the following constructive insights for LfD.

*1) One-Shot LfD of Complex Manipulation Tasks*

We showed that by modeling physical interactions, complex, contact-rich, and sequential tasks can be reproduced using only a single demonstration, conventional hybrid pose/wrench control, and no task-specific prior knowledge (Section V). Other approaches require multiple demonstrations and/or substantial prior knowledge in the form of large pre-trained models that are generally difficult to interpret, debug, extend, provide guarantees for, and require substantial computational resources (Section I). In contrast, our method is interpretable, does not require substantial computational resources, and requires physically meaningful parameters that may be selected for a large range of tasks (Table II). For example, precise assembly tasks may require stricter tracking tolerances and lower velocity thresholds than household tasks.

*2) Robustness to Unknown Geometric Variations*

Whether an LfD method provides sufficient robustness to unknown geometric variations depends on many factors. For example, the task itself, the demonstrator, the environment, the robot, the controller, external systems (e.g. vision), and the measure of task success. Therefore, instead we illustrated *how* robustness to geometric variations may be improved.

First, we reaffirm the need for compliant control in task-relevant reference frames [20], [21], [33]. Hybrid control in task-relevant frames (Section V.B) was essential for several tasks even without geometric variations, and improved robustness in all others. Compliant control inherently provides robustness by keeping interaction wrenches low and guarantees contact where needed. Task-relevant reference frames decouple signals and control goals, thereby simplifying classification, monitoring, and control.

Second, we showed the benefits of explicitly monitoring tracking at the same time as contact transitions (Section V.C).

Tracking within prespecified tolerances guarantees that the required accuracy of sensitive segments is met. For example, when inserting pegs or grasping smaller objects such as cabinet handles. At the same time, contact transitions provide grounded checkpoints in tasks based on online haptic information about the environment, rather than noisy pose estimates, such as from vision. We combined such contact monitoring with sufficiently accurate tracking by relaxing the required tracking tolerances when contact changes were expected. Thereby, increasing the robustness to geometric variations in all experiments.

Third, we illustrated the need to explicitly monitor unilateral contact transitions on translations and rotations. We considered the making of constraints due to making contact and breaking of constraints due to moving off edges or dislodging. We distinguished between constraint transitions that occurred earlier than expected (Section V.C.2.a) or transitions that did not occur, even at the end of the tracked trajectory (Section V.C.2.b). In case the expected transitions did not occur, exploration was needed to find them. Pose exploration by linearly extrapolating the demonstrated reference trajectories was sufficient in most cases, but more elaborate methods are generally needed, such as for two-dimensional hole search (Section VI.D.2). Wrench exploration by increasing the wrench in the direction of an expected constraint loss was sufficient for simple dislodging (Section VI.E.1). More elaborate methods may explore the pose or wrench spaces extensively, or retry an approach by restarting from a previous segment.

Monitoring where contact changes occur, possible after exploration, allows discovery of the previously unknown pose offsets in the environment geometry. This new pose information may be used naturally to adapt the task model for future reproduction. Thereby, robustness obtained through contact awareness leads naturally to adaptivity.

#### 3) *Adaptivity to Online Discovered Geometric Variations*

We showed that adapting task models using new pose and contact information discovered online, possibly after exploration, is often necessary to manage larger geometric variations (Section V.D). For example, if objects are in different locations than during the demonstration, the demonstrated trajectories may need to be adapted, since tracking them within the prespecified tolerances would require mechanical interference.

To adapt task models online, we generalized the trajectories accordingly during reproduction using online haptic information. We used a single sample of the currently measured pose, which provides limited accuracy. To further increase adaptivity, more informative identification is required. For example, by applying our identification method online.

In addition to adapting the task models, adapting the controllers online may also be required in cases where the demonstration itself does not contain enough information. For example, if a demonstration contained no substantial velocities or wrenches, and therefore no perturbations, the desirable response to perturbations (pose or wrench control) cannot be inferred. In this work, we assumed pose control for such axes to avoid drift. Alternatively, the control may be adapted to compliant wrench control if unexpected constraints are measured online.

#### 4) *Generalization using a Priori Known Geometric Variations*

Generalizing task models to new geometric situations explicitly requires that the models contain meaningful parameters that can be changed. We provide two methods (Section IV): applying constant offsets to segments and linking segments with different offsets by blending trajectories. Both methods are commonly necessary. For example, when inserting an object in a new location, where a group of segments has the same offset, followed by polishing another object in another new location, where a different group of segments has another offset.

Vision often provides the first description of a new geometric situation. Our identified task models may provide useful prior information about which visual features are relevant to generalize over. In particular, interaction frames may identify where and on which body task-relevant geometric features are in the demonstration. Examples from our experiments include the hinge latch axis, barrel latch axis, door axis, screw bolt tip, and screwing hole. By estimating the poses of similar geometric features in a new scene, task models from a single demonstration may be generalized to unseen situations, for example with visual semantic correspondence [24]. As long as the objects in a new scene contain corresponding task-relevant geometric features, such as tool tips, other properties such as pose, size, color, and dynamics may differ.

### B. *Limitations and Future work*

To maintain interpretability, we used discrete elements in our task modeling, monitoring, and control based on prespecified thresholds. For example, by modeling discrete interaction classes and choosing between position or force control. Continuous methods may further improve robustness by reducing the sensitivity around the prespecified thresholds, in particular for tasks where the thresholds are hard to choose. Such continuous extensions may include probabilistic classification and monitoring, and smoothly switching the control between segments by changing controller impedances.

Although our work illustrates that tasks *can* be reproduced using a single demonstration, the use of multiple demonstrations provides several advantages. First, multiple demonstrations may average out user errors that otherwise occur in single demonstrations, thereby lowering the requirements on user skill. Second, multiple demonstrations may reveal multiple viable sequences to complete the same task, where one execution path may be more robust than the other depending on the environment. For example, by relying more on contact guiding in situations where pose estimation using vision is error prone. Third, multiple demonstrations may reveal variations in the demonstrated trajectories, which can be used to tune required tracking tolerances or controller stiffness.

Our method relies on several assumptions. We assumed everything that was not grasped by the robot to be part of a static environment. Capturing interactions with moving objects would generally require pose estimation of those moving objects, such as using vision. Furthermore, because we assumed interactions to be rigid (negligible compliance), quasistatic (negligible inertial effects), and without gravity, the impeded motion ($\mathcal{Z}$) interaction class may only be caused by friction. If these assumptions do not hold, it may be necessary to further identify which dynamics caused the impeded motion, to infer suitable control for reproduction.

Controller gains were briefly tuned to result in reasonable reproduction times across all evaluated tasks while maintaining stability. The reference trajectories and wrenches were executed at the same speed as during the demonstration. However, reproduction times were approximately twice as long because sufficient integral control was required to meet the stricter tracking tolerances to switch to the next segment. To reduce reproduction times, the tolerances may be relaxed if they are not the limiting factor. Increasing the control gains to improve tracking may cause instability, in particular for interaction frames that are further from the end effector. Sensor noise has a larger destabilizing effect on the control expressed in the frames that are further away, due to the larger moment arm. Furthermore, the controller gains were limited by our relatively slow 100 Hz Python implementation, in particular for the closed-loop wrench control gains [36]. More efficient control implementations should result in reproduction times closer to the human demonstration. Alternatively, open-loop wrench control may be sufficient for many tasks that require contact but no precise wrench tracking, thereby avoiding the destabilizing effects of wrench feedback.

## Appendix

### C. Modified Oversegmentation Filter

Segmentation using threshold crossings is prone to oversegmentation, because small variations around the thresholds, such as from noise, may add extraneous segments. To reduce oversegmentation, our previous work only accepted threshold crossings for which there were no other threshold crossings in a dead time *window* of 0.3 s before or after [20]. Here, we require a *minimum integral* between threshold crossings, such that the signal magnitudes are also relevant. The filter has three steps (Fig. 9).

First, we only accept threshold crossings for which the absolute integral to the previous or to the subsequent crossing is larger than the product of the dead time and threshold (Fig. 9a). We use the same thresholds $({}^*\omega, {}^*v, {}^*m, {}^*f)$ as for classification. For example, we require the absolute integral of the linear velocity norm $\|\boldsymbol{v}^b(k)\|$ between two changepoints to be larger than $0.3\ \mathrm{s} \cdot 20\ \mathrm{mm\ s^{-1}} = 6\ \mathrm{mm}$. Therefore, a new segment based on the linear velocity can only be made if there has been at least 6 mm of motion.

Second, if there are subsequent crossing pairs that are in the same direction (up-up or down-down), we remove one of the pair (Fig. 9b, bottom). For an up-up pair we remove the first crossing if the integral is negative, and the second crossing if the integral is positive. For a down-down pair, the logic is opposite. Hence, ambiguous segments are combined with the one before or after.

Third, we remove up-down or down-up pairs if the absolute integral between them is below the minimum integral, to remove short segments due to momentary spikes (Fig. 9b).

Lastly, we pool the separate threshold crossings, and average them in time if they are within the dead time window (Fig. 9b).

(a)

(b)

**Fig. 9.** Example of segmenting a simple task, consisting of idle waiting (**1**), moving in free space (**2**) until contact is made (**3**), and relieving the contact force (**4**).